\documentclass[twoside]{article}

\usepackage[preprint]{aistats2023}
\usepackage[round]{natbib}
\renewcommand{\bibname}{References}
\renewcommand{\bibsection}{\subsubsection*{\bibname}}

\usepackage{amsfonts,bm}
\usepackage{hyperref}
\usepackage{booktabs}

\usepackage{xcolor}
\usepackage{amsmath}
\usepackage{graphicx}
\usepackage{subfiles}
\usepackage{amsthm}
\usepackage{thmtools} 
\usepackage{thm-restate}
\usepackage{tikz}
\usetikzlibrary{shapes,arrows}
\usetikzlibrary{arrows,calc,positioning}
\tikzset{
    block/.style = {draw, rectangle,
        minimum height=1cm,
        minimum width=2cm},
    input/.style = {coordinate,node distance=1cm},
    output/.style = {coordinate,node distance=4cm},
    arrow/.style={draw, -latex,node distance=2cm},
    pinstyle/.style = {pin edge={latex-, black,node distance=2cm}},
    sum/.style = {draw, circle, node distance=1cm},
}
\usetikzlibrary{bayesnet}
\usepackage{algorithm}
\usepackage{algpseudocode}
\usepackage{pbox}
\usepackage{array}
\usepackage{makecell}
\usepackage{wrapfig}
\usepackage{soul}
\usepackage{mdwlist}

\DeclareMathOperator*{\argmin}{\arg\!\min}
\DeclareMathOperator*{\argmax}{\arg\!\max}
\DeclareMathOperator{\ReLU}{ReLU}
\DeclareMathOperator{\tReLU}{ReLU_\tau}
\DeclareMathOperator{\LtReLU}{LReLU_\tau}
\DeclareMathOperator{\expfam}{\mathcal{D}_{A,\chi}}
\newcommand{\indep}{\perp \!\!\! \perp}

\newcommand{\Rname}{canonical nonlinearity}

\newcommand\numberthis{\addtocounter{equation}{1}\tag{\theequation}}

\declaretheorem[name=Lemma]{lemma}

\declaretheorem[name=Assumption]{assumption}

\DeclareRobustCommand{\cheng}[1]{\marginpar{\textcolor{purple}{Cheng:~{#1}}}}
\DeclareRobustCommand{\russ}[1]{\marginpar{\textcolor{green}{Russ:~{#1}}}}

\begin{document}

% If your paper is accepted and the title of your paper is very long,
% the style will print as headings an error message. Use the following
% command to supply a shorter title of your paper so that it can be
% used as headings.
%
%\runningtitle{I use this title instead because the last one was very long}

% If your paper is accepted and the number of authors is large, the
% style will print as headings an error message. Use the following
% command to supply a shorter version of the authors names so that
% they can be used as headings (for example, use only the surnames)
%
%\runningauthor{Surname 1, Surname 2, Surname 3, ...., Surname n}

\twocolumn[

\aistatstitle{Deep equilibrium models as estimators for continuous latent variables}

\aistatsauthor{ Russell Tsuchida \And Cheng Soon Ong }

\aistatsaddress{ Data61, CSIRO, Australia \And  Data61, CSIRO \\ and Australian National University} ]

%\russ{Changes for NeurIPS submission: The extended table of experiments is moved from Appendix to main paper. Changed the wording of Theorem 1. Added some figures concerning $\tReLU$ to the main paper. I have changed the story in the abstract, focusing more on the connection that we make rather than the method that we propose. The method is actually just what people would use already --- the contribution is that now there is a reason for using this method beyond black-box performance.}
%\cheng{I shortened the abstract, hope we didn't lose anything important.}

\begin{abstract}
Principal Component Analysis (PCA) and its exponential family extensions have three components: observations, latents and parameters of a linear transformation. 
We consider a generalised setting where the canonical parameters of the exponential family are a nonlinear transformation of the latents. 
We show explicit relationships between particular neural network architectures and the corresponding statistical models.
We find that deep equilibrium models --- a recently introduced class of implicit neural networks --- solve maximum a-posteriori (MAP) estimates for the latents and parameters of the transformation.
Our analysis provides a systematic way to relate activation functions, dropout, and layer structure, to statistical assumptions about the observations, thus providing foundational principles for unsupervised DEQs.
For hierarchical latents, individual neurons can be interpreted as nodes in a deep graphical model. Our DEQ feature maps are end-to-end differentiable, enabling fine-tuning for downstream tasks.
%Principal Component Analysis (PCA) and its exponential family extensions have three components: observed variables, latent variables and parameters of a linear transformation. The likelihood of the observation is an exponential family with canonical parameters that are a linear transformation of the latent variables. We consider a generalised setting where the canonical parameters are a nonlinear transformation of the latent variables, which we call nonlinearly parameterised exponential family PCA. We propose a flexible computational method to project the observation onto latent space, that is, to infer the value of latent variables. Joint maximum a-posteriori (MAP) estimates can be computed using a deep equilibrium model, which we call Principal Equilibrium Dimensions (PED). PED provides a systematic way to relate choices of neural network activation functions, dropout and layer structure back to statistical assumptions about the observations. PED layers can be guaranteed to admit unique solutions. Finally, we extend the setting to the case of hierarchical latent variables, via deep PED. An advantage of PED is that it is end-to-end differentiable, and can be fine-tuned in downstream tasks.
\end{abstract}
% The likelihood of the observation is an exponential family with canonical parameters that are a linear transformation of the latent variables. 

\section{Introduction}

\looseness=-1  %This causes LaTeX to shrink a paragraph with a one word orphan

Deep learning provides a means of fitting flexible function classes to data. 
Function classes are defined as compositions of parameterised functions called layers.
Parameters of the layers are typically jointly adjusted by applying first-order optimisation methods to an objective involving the data and the parameters.
%~\citep[\S 5.6.1]{deisenroth20mathml}. 
Layers may provide an explicit description of a mapping. 
For example, given an input $x \in \mathbb{R}^D$, the $L$-dimensional output of a layer with parameters $\theta \in \mathbb{R}^{L \times D}$ might be defined by $\sigma\big( \theta x\big)$.
%for some function $\sigma$ applied element-wise. 

Alternatively, layers may be defined implicitly as solutions to certain problems.
For example, given an input $x \in \mathbb{R}^D$, a Deep Equilibrium Model (DEQ)~\citep{bai2019deep} layer might output a solution to the fixed point equation $z = \sigma\big( \gamma z + \theta x\big)$, where $\theta \in \mathbb{R}^{L \times D}$ and $\gamma \in \mathbb{R}^{L \times L}$ are parameters and $z \in \mathbb{R}^{L}$ is the implicitly defined output.
Implicit layers include DEQs which solve fixed point equations, Neural ODEs~\citep{chen2018neural} which solve ODEs and Deep Declarative Networks (DDNs)~\citep{gould2021deep} which solve optimisation problems. All implicit layers carry an issue of well-posedness, that is, whether there exists a unique solution to the problem that defines the layer.
Implicit networks' parameters retain the ability of being easy to adjust using first-order optimisation methods. 
Under mild conditions, the gradient of the layer can be computed without back-propagating through the solver that computes the output of the implicit layer.

The use of DEQs, and deep learning models more generally, is usually motivated from a top-down view. One uses a deep learning model for some combination of predictive performance (measured empirically \emph{a posteriori}), or representational capacity (proven mathematically \emph{a priori}). In contrast, classical statistical models can often be motivated from a bottom-up data generating process and are therefore interpretable~\citep{rudin2019stop}, but might not necessarily achieve as good performance for a given task. Our paper closes this gap, by understanding deep learning models through data generating processes~\citep{efron2020prediction}. 

\begin{figure*}
    \centering
    \begin{minipage}{.45\textwidth}
        \centering
            \begin{tikzpicture}
              % Define nodes
              \node[obs]                                (y) {$Y_s$};
              \node[latent, above=of y, xshift=0.5cm]   (w) {$\mathsf{W}$};
                \node[latent, above=of y, xshift=-0.5cm]   (b) {$B$};
              \node[latent, right=of y]                  (z) {$Z_s$};
            
              % Connect the nodes
              \edge {z,w,b} {y} ; %
            
              % Plates
              \plate[inner sep=0.35cm] {yz} {(z)(y)} {$s=1,\ldots,N$} ;
            \end{tikzpicture} 
        \end{minipage}%
    \begin{minipage}{0.55\textwidth}
    \centering
            \begin{tikzpicture}[auto, node distance=1cm,>=latex']
            \node [input, name=input] {$Y_s$};
            \node [align=center, block, right=2cm of input] (controller) {DEQ Layer \\ $Z_s^\ast = f(Z_s^\ast; Y_s, \mathsf{W}, B)$};
            \node [output, right=2cm of controller] (output) {};
            \node at ($(controller)!0.5!(controller)+(0,-2)$) [align=center, block] (feedback) {Loss \\ $-\log p\big(Z_s^\ast, \mathsf{W}, B \mid Y_s \big)$};
            \draw [draw,->] (input) -- node [name=branch] {$Y_s$} (controller);
            \draw [->] (controller) -- node [name=y] {$Z_s^\ast$}(output);
            \draw [->] (y) |- (feedback) ;
            \draw [<-] (feedback) -| (branch);
            \draw [draw,->] (feedback) -- node[pos=0.2] {\tiny{backprop}} (controller);
            \end{tikzpicture}   
        \end{minipage}
    \caption{(Left) The graphical model for nonlinearly parameterised exponential family PCA with observations $Y_s$, latents $Z_s$, and parameters $\mathsf{W}, B$. The likelihood of the observation $Y_s$ is an exponential family with a canonical parameter $R(\mathsf{W} Z_s + B)$, for some nonlinearity $R$. (Right) MAP estimates can be found using PED. Latents are estimated by the prediction of the DEQ layer given fixed parameters. Parameters of the DEQ layer are adjusted through backpropagation.}
    \label{fig:bayes_nets}
\end{figure*}
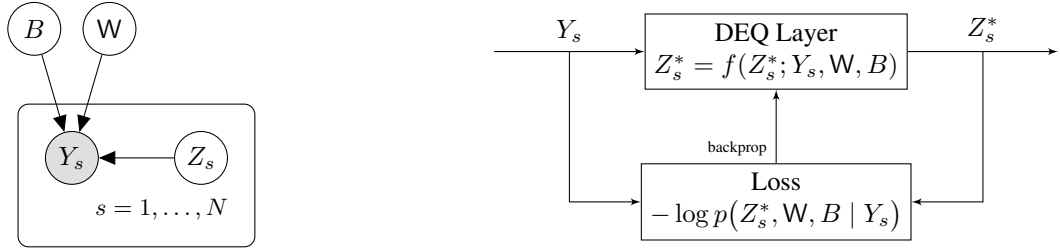

\paragraph{Contributions.} We present Principal Equilibrium Dimensions (PED), a DEQ that solves the problem of joint MAP estimation in a graphical model representing nonlinearly parameterised exponential family PCA. We stress that our contributions are mostly theoretical in nature, with the aim of closing the gap between black box predictors and data generating models. Our analysis provides a bottom-up justification for the use of particular DEQ layers in unsupervised learning. PED layers have similar components to explicit networks, such as activation functions and dropout. PED layers are end-to-end differentiable and may be fine-tuned in supervised settings. Our contributions are:
\vspace{-0.5cm}
\begin{itemize*}
    \item We show that MAP estimates of the latents and parameters of nonlinearly parameterised exponential family PCA can be computed by a DEQ layer. See Figure~\ref{fig:bayes_nets}.
    \item  We relate statistical assumptions about the data to a choice of activation function, giving conceptual meaning to the use of activations such as $\tanh$, logistic sigmoid, softmax, and $\ReLU$,
    as well as a form of learned dropout. See Table~\ref{tab:log-partition}.
    \item In the nicest setting (Theorem~\ref{thm:general_shallow}), PED layers compute solutions to strongly convex optimisation problems. Hence they are guaranteed to admit unique fixed points. In less nice settings (Theorem~\ref{thm:relu_shallow}), we ensure that PED layers compute local minima.
    \item We derive a principled method for constructing a DEQ that solves a graphical model with hierarchical latent variables, called deep PED. See Figure~\ref{fig:bayes_nets_deep}.
    \item We provide an implementation for shallow and deep PED and compare it to PCA, tSNE and UMAP on illustrative synthetic datasets (Figure~\ref{fig:synthetic}).
\end{itemize*}

%In contrast with~\citet{tsuchida2021declarative} where weights are constrained to be symmetric positive semi definite, our derived model optimises over an unconstrained set of $d \times l$ weight parameter matrices and biases.
\section{Background and notation} 
\label{sec:background}
\paragraph{Exponential families.} Recall the definition of an exponential family, which is a class of
probability distributions whose sufficient statistics admit a moment generating function defined on an open set $\mathbb{F} \subseteq \mathbb{R}$. For canonical parameters $v$, define the \emph{log partition function} $A:\mathbb{F}\to\mathbb{R}$. The probability density function (or probability mass function)
\begin{align*}
    p(y \mid {v}) = h(y) \exp\big( {v} T(y) - A({v}) \big) \numberthis \label{eq:exp_fam}
\end{align*}
represents a minimal regular exponential family parameterised by $v$. For such a family, $A$ is both infinitely differentiable and strictly convex~\citep[Proposition 3.1]{wainwright2008}. See Appendix~\ref{sec:expfam-construction} for more details.

$A$ acts as a cumulant generating function, in particular the expected value under $p(y \mid v)$ is $\mu = A'(v)$. The parameter $\mu$ is called the expectation parameter.
%Following common convention, we use $p( \cdot \mid \cdot )$ to denote all (conditional) probability density functions, with a meaning clear from the arguments. 
We write $y \sim \mathcal{D}_{A,\chi}(v)$ to mean that a scalar-valued random variable $y$ follows an exponential family with log partition function $A$, scalar canonical parameter $v$ and additional parameter $\chi$\footnote{Some exponential families involve a parameter that is not canonical. E.g. the univariate Gaussian with known variance has a scale parameter~\citep[p. 16]{nielsen2009statistical}, and Laplace has a centering parameter. Both $A$ and $\chi$ are fixed.}. If $Y=(y_1, \ldots, y_d)$ and $V = (v_1, \ldots, v_d)$, we write $Y \sim \mathcal{D}_{A,\chi}(V)$ to mean $y_i \sim \mathcal{D}_{A,\chi}(v_i)$ for all $1 \leq i \leq d$, with each $y_i$ mutually conditionally independent given $V$.

%\cheng{We need a notation convention description. E.g. $\mathsf{Y}$ is a matrix corresponding to collecting the column vectors $Y_s$. An element of a $d$ dimensional vector $Y_s$ is denoted $y_{si}$ where $i=1,\ldots,d$,}  % This is in the last paragraph of section 2. Does it help?
\paragraph{Matrices and data.} We use $\mathsf{I}$ to denote a square conformable identity matrix, with dimensions defined from context. Let $Y_s=(y_{s1}; \ldots; y_{sd}) \sim \mathcal{D}_{A,\chi}(V_s)$, where for each $s$, $V_s \in \mathbb{R}^{d}$ is a vector consisting of canonical parameters.
Let $\mathsf{Y} \in \mathbb{R}^{d \times N}$ denote a matrix with $N$ such vectors $Y_s$, $1 \leq s \leq N$, each sampled conditionally independently given the canonical parameters. Each of the $Y_s$ are associated with a latent representation $Z_s \in \mathbb{R}^l$. Write $\mathsf{Z} \in \mathbb{R}^{l \times N}$ for the matrix consisting of such latent representations.
We sometimes drop subscripts for cleanliness when they are fixed. 
% Cheng: We could put this back in if we have the space.
%Scalars are lower-case, vectors are upper-case and matrices are upper-case sans-serif. Functions that accept scalars extend to vector and matrix inputs by returning the function evaluated elementwise.
Functions are composed using $\circ$, and $\odot$ represents elementwise product of vectors.

\begin{table*}[!ht]
    \centering
\begin{tabular}{cc|c|c|c}
&                                                             $R(\eta)$                       & $A \circ R \, (\eta) $                               & $\rho (\eta) := R'(\eta)$                         & $\sigma(\eta) := (A \circ R)' \, (\eta)$                              \\  \toprule \hline
\multicolumn{5}{|l|}{Linear $R$, Quadratic $A$ \textcolor{gray}{(Classical PCA)}}                                                                                                                                                                                                                                                                                \\ \hline
&  \multicolumn{1}{c}{$\eta$}         & \multicolumn{1}{c}{${\eta}^2/2$}                       & \multicolumn{1}{c}{$1$}               & \multicolumn{1}{c}{${\eta}$}                                            \\ \hline
\multicolumn{5}{|l|} {Linear $R$, General $A$ \textcolor{gray}{(Exponential family PCA)}}                                                                                                                                                                                                                                                                                  \\ \hline
& \multicolumn{1}{c}{$\eta$}         & \multicolumn{1}{c}{$\log\big(1+\exp({\eta})\big)$}     & \multicolumn{1}{c}{$1$}               & \multicolumn{1}{c}{$\frac{\exp({\eta})}{\exp({\eta})+1}$}                 \\
& \multicolumn{1}{c}{$\eta$}         & \multicolumn{1}{c}{$\log \cosh {\eta}$}                & \multicolumn{1}{c}{$1$}               & \multicolumn{1}{c}{$\tanh({\eta})$}                                    \\
& \multicolumn{1}{c}{$\eta$}         & \multicolumn{1}{c}{$\log \frac{e^{\eta} - 1}{{\eta}}$} & \multicolumn{1}{c}{$1$}               &\multicolumn{1}{c}{$\frac{e^{\eta}({\eta}-1)+1}{(e^{\eta}-1){\eta}}$}   \\ \hline
\multicolumn{5}{|l|}{Nonlinear $R$, Quadratic $A$ \textcolor{gray}{(Non-linearly parameterised classical PCA)}}                                                                                                                                                                                                                                                                             \\ \hline
 & \multicolumn{1}{c}{$\tReLU(\eta)$} & \multicolumn{1}{c}{$\tReLU^2(\eta)/2$}                 & \multicolumn{1}{c}{$\Phi_\tau(\eta)$} & \multicolumn{1}{c}{$\Phi_\tau(\eta) \tReLU(\eta)$}                     \\
 & \multicolumn{1}{c}{$\ReLU(\eta)$}  & \multicolumn{1}{c}{$\ReLU^2({\eta})/2$}                & \multicolumn{1}{c}{$\Theta(\eta)$}    & \multicolumn{1}{c}{$\ReLU({\eta})$}                                    \\ 
 & \multicolumn{1}{c}{$\eta + \log \cosh \eta$}  & \multicolumn{1}{c}{$(\eta + \log \cosh \eta)^2/2$}                & \multicolumn{1}{c}{$1 + \tanh(\eta)$}    & \multicolumn{1}{c}{$(\tanh(\eta)+1)(\eta+ \log \cosh \eta) $}                                    \\  \hline
\multicolumn{5}{|l|}{Nonlinear $R$, General $A$ \textcolor{gray}{(Non-linearly parameterised exponential family PCA)}}           
\\ \hline
 & \multicolumn{1}{c}{$-\tReLU(\eta)$} & \multicolumn{1}{c}{$-\log \tReLU(\eta)$}                 & \multicolumn{1}{c}{$-\Phi_\tau(\eta)$} & \multicolumn{1}{c}{$-\frac{\Phi_\tau(\eta)}{\tReLU(\eta)}$}                     \\
 & \multicolumn{1}{c}{$\tReLU(\eta)$} & \multicolumn{1}{c}{$t \log\Big(1 + \exp \big(\tReLU(\eta) \big) \Big)$}                 & \multicolumn{1}{c}{$\Phi_\tau(\eta)$} & \multicolumn{1}{c}{$t \frac{\exp \big(\tReLU(\eta) \big)}{1 + \exp \big(\tReLU(\eta) \big)} \Phi_\tau(\eta)$}                     \\
 & \multicolumn{1}{c}{$\ReLU(\eta)$} & \multicolumn{1}{c}{$t \log\Big(1 + \exp \big(\ReLU(\eta) \big) \Big)$}                 & \multicolumn{1}{c}{$\Theta(\eta)$} & \multicolumn{1}{c}{$t \frac{\exp \big(\ReLU(\eta) \big)}{1 + \exp \big(\ReLU(\eta) \big)} \Theta(\eta)$}               \\
& \multicolumn{1}{c}{$-\tReLU$}         & \multicolumn{1}{c}{$\text{Li}_{j+1}\big( \exp( -\tReLU(\eta)) \big)$}        & \multicolumn{1}{c}{$-\Phi_\tau(\eta)$}               & \multicolumn{1}{c}{$-\text{Li}_{j}\big( \exp( -\tReLU(\eta)) \big) \Phi_\tau(\eta)$}                                    \\ \bottomrule
\end{tabular}
    \caption{\looseness=-1 Different combinations of $A$ and $R$ induce different soft dropout and activation functions $\rho$  and $\sigma$. The first four rows represent cases of exponential family PCA with Gaussian, Bernoulli, non-interacting Ising and continuous Bernoulli~\citep{loaiza2019continuous} likelihoods respectively. Our setting allows for general $A$ and $R$, all of which can be solved using single-layer PED. Of particular interest is the case where $A$ is quadratic but $R$ is general, where an extended setting involving a hierarchy of latent variables can be solved using deep PED. Further details of the probabilistic interpretation of each of these cases is described in Appendix~\ref{app:log_partition}. The function $\tReLU$ is discussed in Appendix~\ref{app:relu}. The functions $\Phi_\tau$ and $\Theta$ are CDFs of a Gaussian with standard deviation $|\tau|$ and the Heaviside step function respectively. The function $\text{Li}_{j+1}$ is the polylogarithm of order $j+1$. See also~\cite{nielsen2009statistical} for the case $\rho = 1$.}
    \label{tab:log-partition}
\end{table*}

\paragraph{Implicit neural networks.} Our analysis will reveal that MAP estimates of nonlinearly parameterised exponential family PCA can be computed by training a DEQ in an unsupervised setting. Connections between DEQs and DDNs (which solve optimisation problems) have previously been explored by~\citet{tsuchida2021declarative,li2021optimization}. We work with a prototype problem to represent a DEQ,
\begin{align*}
    \min_{\theta, \gamma} \quad &\sum_{s=1}^N \mathcal{L}\Big(\mathcal{F}_\gamma\big(Z_s^\ast, \, Y_s \big), \, Y_s \Big) \numberthis \label{eq:implicit_net} \\
    \text{subject to } \quad &Z_s^\ast = g(\theta, Y_s, Z_s^\ast), \quad s = 1, \ldots, N. 
\end{align*}
\looseness=-1 Here $\theta$ are parameters of the implicit layer, $Z_s^\ast$ are outputs, $\mathcal{F}_\gamma$ is a neural network (that may also contain implicit layers) parameterised by $\gamma$, and $\mathcal{L}$ is a loss function. 
The constraint is satisfied at a fixed point of a function $g$, which depends on parameters $\theta$ and input data $Y_s$.

\looseness=-1 The outer problem learns the parameters of the network, and the inner constraint outputs predictions of the implicit layer.
The inner constraint is subject to questions of well-posedness, that is, whether solutions to the problem exist and are unique.
The workhorse tools for analysing well-posedness are contraction principles for fixed point equations~\citep{hasselblatt2003first} and convexity for optimisation problems~\citep[Chapter 2]{wright_recht_2022}. 
Following usual deep learning philosophy, we ignore such questions for the outer parameter learning problem, which is likely to be highly nonconvex.
Our aim is to cast problems in the form~\eqref{eq:implicit_net} with well-posed inner constraints. A problem cast in the form of~\eqref{eq:implicit_net} can use the machinery of implicit deep learning --- the implicit function theorem and GPU-optimised libraries --- to solve the problem.

\paragraph{Variants of PCA.} PCA admits linear algebraic, probabilistic~\citep{bishop2006pattern} and functional interpretations~\citep{scholkopf1997kernel}. We work with the probabilistic view, since it allows for extensions with intuitive statistical meaning. Variants of PCA are reviewed by~\citet{smallman2022literature}. 

%The main reason to have a dash is when it is unclear which word is the describer, and which word is described (usually when there are three words). And so a-posteriori makes sense, but log-partition doesn't. It really should be log partition-function. It is not the "log partition" of a function. It is the "log" of the "partition function". If you end up with sentences where there is potential confusion, which happens when you have two compound phrases in the same sentence, I recommend to rephrase those sentences using prepositions. E.g. log of the partition function.
\section{Derivation of PED}
\label{sec:shallow}
We propose to insert nonlinear mappings $R:\mathbb{R} \to \mathbb{F}$ called the \emph{\Rname} into the exponential family. In a sense to be made more precise, we work with a composition $A \circ  R (\cdot)$. Various combinations of $A$ and $R$ are given in Table~\ref{tab:log-partition}.
The derivatives $\rho := R'$ and $ \sigma := (A \circ R)'$ (where defined) play important roles, and we call them the \emph{soft dropout} and \emph{activation} functions respectively.
We consider the graphical model in Figure~\ref{fig:bayes_nets}, which mirrors various versions of PCA~\citep{bishop2006pattern,collins2001generalization,mohamed2008bayesian} but allows non-identity $R$.

\subsection{MAP estimation using DEQs}
\paragraph{Setup.}
Let $\mathsf{H} = \mathsf{W} \mathsf{Z} + B 1_{1 \times N}$, where $\mathsf{W} \in \mathbb{R}^{d \times l}$, $B \in \mathbb{R}^d$ and $\mathsf{Z} \in \mathbb{R}^{l \times N}$. 
Equivalently $H_s = \mathsf{W} Z_s + B$ where $Z_s \in \mathbb{R}^{l \times 1}$ are the latent variables for data $Y_s=(y_{s1}, \ldots y_{sd})$.
Typically $l < d$, although this need not be the case.
We introduce a (potentially nonlinear) canonical map $R(\cdot)$.
The canonical parameters of the distribution of the observed variables $Y_s$ are $R(H_s) \in \mathbb{F}$. 
In other words, we assume that the observed vector of data $Y_s$ is drawn from a nonlinear parameterisation of the
exponential family $\mathcal{D}_{A, \chi}(R(H_s))$ with log partition function $A$, and additional parameter $\chi$.
This immediately implies that the distribution is properly normalised.
The nonlinear parameterisation does not violate the Fisher-Neyman factorisation (see~\citet[Theorem 6.14]{deisenroth20mathml}). That is, $T$ is still a sufficient statistic for $\mathsf{W}Z_s + B$. 
The expectation parameter satisfies $\mu = \sigma(\eta)/\rho(\eta)$ when the right hand side is defined (see Appendix~\ref{app:bregman}).
Following \eqref{eq:exp_fam}, our proposed nonlinear exponential family is
%Equivalently $H_s = \mathsf{W} Z_s + B$ where $Z_s \in \mathbb{R}^{l \times 1}$ are the latent variables for data $Y_s=(y_{s1}, \ldots y_{sd}) \sim \mathcal{D}_{A, \chi}(H_s)$. 
%We suppose that the canonical parameters for $\mathsf{Y}$ are given by $R(\mathsf{H})$, or equivalently that the canonical parameters for $Y_s$ are given by $R(Y_s)$.
\begin{align*}
&\phantom{{}={}} p( \mathsf{Y} \mid \mathsf{Z}, \mathsf{W}, B) = \prod_{s=1}^N \prod_{i=1}^d h\big( y_{si} \big) \numberthis \label{eq:nonlinear-expfam} \\
&  \exp \Big( R \big((\mathsf{W} Z_s + B)_i \big) T(y_{si})  - A \circ R \, \big( (\mathsf{W} Z_s + B)_i \big) \Big).
\end{align*}
For each $s$, place independent zero mean iid Gaussian priors over $Z_s$, $Z_s \sim \mathcal{N}(0, \lambda^{-1}\mathsf{I}), {Z_1 \indep \ldots \indep Z_N}$.

There are two settings of interest. In the first, the parameters $\mathsf{W}$ and $\mathsf{B}$ are fixed and belong to a nonempty set $\mathbb{W} \subseteq \mathbb{R}^{d \times l + d}$. The posterior over $\mathsf{Z}$ satisfies
\begin{align*}
    p\big(\mathsf{Z} \mid \mathsf{W}, B, \mathsf{Y} \big)&= p\big( \mathsf{Y} \mid \mathsf{Z}, \mathsf{W}, B \big) p(\mathsf{Z}) / p(\mathsf{Y} \mid \mathsf{W}, B). \numberthis \label{eq:posterior_Z}
\end{align*}
In the second setting, we place a prior $\mathcal{P}$ admitting a density supported on $\mathbb{W}$ over $\mathsf{W}$ and $B$. The joint posterior is
\begin{align*}
    p( \mathsf{Z}, \mathsf{W}, B \mid \mathsf{Y}) &= p( \mathsf{Y} \mid \mathsf{Z}, \mathsf{W}, B) p(\mathsf{Z}) p(\mathsf{W}, B)/p(\mathsf{Y}).\numberthis \label{eq:posterior}
\end{align*}
In Appendix~\ref{app:gaussian}, we show the special case where the likelihood is Gaussian and $R$ is the identity, i.e. probabilistic PCA. This special case recovers the usual MAP estimate for $\mathsf{Z}$, as well as an auto-encoder style reprojection error objective for the parameters $\mathsf{W}$ and $B$. 

\paragraph{Results.} We now turn to the more general case of nonlinearly parameterised exponential family likelihoods~\eqref{eq:nonlinear-expfam}. Our conditions for well-posedness of PED layers can be understood from the perspective of optimisation or fixed points. Both views lead to a spectral constraint on $\mathsf{W}$ up to a constant $\kappa$ which includes affects from the observation $\mathsf{Y}$, log partition function $A$ and \Rname $R$.
\begin{assumption}
\label{ass:w_set}
The support $\mathbb{W}$ satisfies
\begin{align*}
\mathbb{W} \subseteq \Big\{ \mathsf{W} \in \mathbb{R}^{d \times l}, B \in \mathbb{R}^d \mid  \kappa \Vert \mathsf{W}^\top \mathsf{W} \Vert_2 < 1 \Big\}, \end{align*}
where $\kappa = \lambda^{-1} \Big( \sup \limits_{\eta, s, i}  T(y_{si}) \rho'(\eta) - \sigma'(\eta) \Big).$

\end{assumption}
\looseness=-1 Note that if $R$ is the identity, $\mathbb{W}$ may be \emph{any} subset of $\mathbb{R}^{d \times l + d}$ since $\rho' = 0$ and $\sigma'$ is nonnegative by convexity of $A$.  
Assumption~\ref{ass:w_set} ensures well-posedness when $R$ is the identity or its second derivative is still relatively well-behaved. However, cases of interest such as $\ReLU$ require a different analysis (see \S~\ref{sec:relu}).
Under Assumption~\ref{ass:w_set}, the first order optimality conditions are sufficient to characterise the unique global optima of the latents given a fixed set of parameters. This first order optimality condition can be rearranged to a fixed point equation, which yields the following DEQ of the form~\eqref{eq:implicit_net}.
\begin{restatable}{theorem}{generalshallow}
\label{thm:general_shallow}
\looseness=-1  Suppose Assumption~\ref{ass:w_set} holds. Let 
\begin{align*}
&\phantom{{}={}} f(Z;Y_s,\mathsf{W},B) \\
&= \frac{1}{\lambda} \mathsf{W}^\top \big( T(Y_s) \odot \rho(\mathsf{W}Z + B) - \sigma(\mathsf{W} Z + B) \big). \numberthis \label{eq:layer}
\end{align*}
Any MAP estimate of~\eqref{eq:posterior_Z} satisfies 
\begin{align*}  Z_s^\ast = f(Z_s^\ast; Y_s, \mathsf{W}, B), \quad \forall 1 \leq s \leq N,  \numberthis \label{eq:fixed_point_layer}
\end{align*}
solutions of which are guaranteed to exist and be unique. Any joint MAP estimate of~\eqref{eq:posterior} satisfies 
\begin{align*}
    &\mathsf{W}^\ast,B^\ast \in \argmin_{\mathsf{W}, B \in \mathbb{W}} - \log p(\mathsf{W},B) - \log p(\mathsf{Z}^\ast) + \\ 
    &\phantom{{}={}} \sum_{s=1}^N 1_{d \times 1}^\top A \circ R \,  (\mathsf{W} Z_s^\ast + B) - T(Y_s)^\top R(\mathsf{W} Z_s^\ast + B) \\
    &\text{subject to } Z_s^\ast = f(Z_s^\ast; Y_s, \mathsf{W}, B), \quad \forall 1 \leq s \leq N.
\end{align*}
\end{restatable}
The proof is given in Appendix~\ref{app:proofs}.
Our analysis relies on the quadratic negative log Gaussian prior $\log p(z)$, which when differentiated yields $\lambda z$. When this term is isolated, a fixed point condition is obtained. When we extend our analysis to hierachical latent variables, the prior $p(z)$ will be replaced by a layer-dependent likelihood, for example $p(Z^{(1)} \mid Z^{(2)})$. This means that $A$ for layers $l > 1$ must be quadratic, but crucially, $R$ may be nonlinear.

The fixed point~\eqref{eq:fixed_point_layer} of a fully connected or convolutional layer~\eqref{eq:layer} represents a DEQ with activation $\sigma$, soft dropout $\rho$ shared weights or convolutional kernel (see Appendix~\ref{app:conv}) $\mathsf{W}$ and biases $B$. We name $\rho$ soft dropout because (a) as shown in Table~\ref{tab:log-partition}, it is often close to a $\{0,1\}$-valued function, and (b) it's role is to multiply inputs $T(Y_s)$ by a value that is close to zero or one. This resembles common usage of dropout in explicit networks --- we are not aware of any existing approaches to incorporating dropout into DEQs.
While we are not aware of any usage of DEQs applied to dimensionality reduction,~\eqref{eq:layer} and~\eqref{eq:fixed_point_layer} resemble what practitioners might use as a generic black-box DEQ  predictor, without any explicit intention to compute with an underlying statistical model. The outer optimisation is a non-Gaussian exponential family generalisation of a squared reprojection error, as detailed in Appendix~\ref{app:gaussian}.

\iffalse
\paragraph{Inference is not coordinate-free.} Unfortunately, our choice of parameterisation effects the space of priors that are easy to work with, and different choices of priors lead to different MAP estimates. 
There are methods for constructing priors that are invariant with respect to coordinate systems~\citep{jeffreys1946invariant}, but these are not considered. This limitaton is not unique to our setting~\citep{bishop2006pattern,collins2001generalization,mohamed2008bayesian}. To illustrate this issue, we sketch a variant where an orthonormal basis is used in place of the unconstrained matrix $\mathsf{W}$ in Appendix~\ref{app:orthonormal}. This variant can also be solved with a DEQ, but has two parameter matrices, one which is a polar decomposition of the other. We leave the practical implications of the choice of coordinate system an open question for future work.
\fi

\subsection{Rectified activations and hard dropout}
\label{sec:relu}
%\subfile{../figures/bayesnet_relu}
\paragraph{Smooth ReLU.}
Define the ReLU by $\ReLU(\eta) = \Theta(\eta) \eta $, where $\Theta(\cdot)$ is the Heaviside function. Let $\tReLU(\eta) = \int_\mathbb{R} \ReLU(\eta + \epsilon) p_\tau (\epsilon) \, d\epsilon$ denote the $\tau$-smoothed ReLU. Here $p_\tau$ is the PDF of a zero-mean Gaussian random variable with standard deviation $|\tau|$. We have
\begin{align*}
    \tReLU(\eta) &= \eta \Phi\Big(\frac{\eta}{|\tau|} \Big) + \frac{|\tau|}{2} \sqrt{\frac{2}{\pi}} \exp \big(-\eta^2/2\tau^2 \big) \\
        \frac{\partial }{\partial \eta} \tReLU(\eta) &= \Phi(\eta/|\tau|)
\end{align*}
\begin{align*}
    \frac{\partial^2 }{\partial \eta^2} \tReLU(\eta) &=  \frac{1}{\sqrt{2 \pi \tau^2}} \exp\big( -\frac{\eta^2}{2\tau^2}\big) = p_\tau(\eta),
\end{align*}
where $\Phi$ is the cdf of the standard Gaussian random variable. %Furthermore, $\tReLU$ converges uniformly to $\ReLU$ as $\tau \to 0$. 
See Appendix~\ref{app:relu} for details and Figure~\ref{fig:relu_activation} for a visualisation of $\sigma$ and $\rho$ when $A(r)=r^2/2$.

\begin{figure}
    \centering
    \includegraphics[scale=0.25]{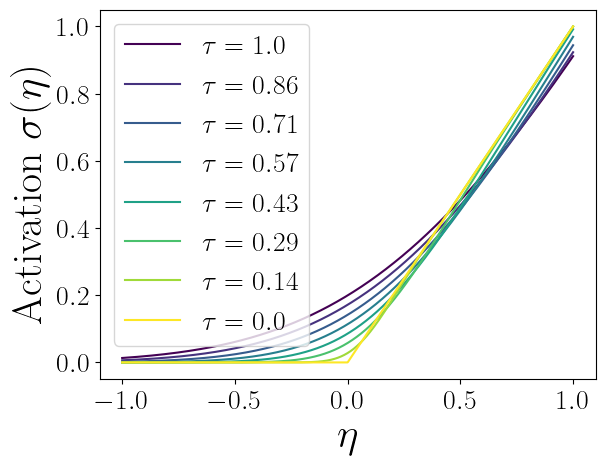}
    \includegraphics[scale=0.25]{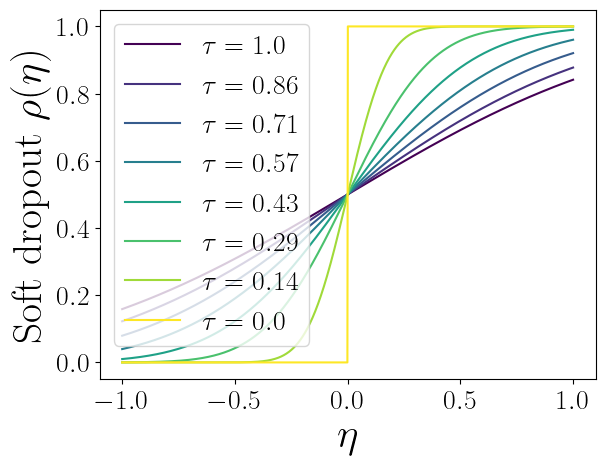}
    \caption{ Choosing $A(r)=r^2/2$ and $R=\tReLU$, (Left) Activation function $\sigma(\eta) = \Phi_\tau(\eta) \tReLU(\eta)$ (Right) Soft dropout function $\rho (\eta) = \Phi_\tau(\eta)$ for different values of $\tau$.}
    \label{fig:relu_activation}
\end{figure}

\paragraph{The need for a separate analysis.}
We consider the special case of Theorem~\ref{thm:general_shallow} when the exponential family is Gaussian and $R = \tReLU$. 
We find $\rho(\eta) = \Phi_\tau(\eta)$ and $\sigma(\eta) = \Phi_\tau(\eta) \tReLU(\eta)$. When $\tau$ is small, $\rho$ and $\sigma$ begin to look like the Heaviside step function and the $\ReLU$. This amounts to using a DEQ consisting of dropout and $\ReLU$ activations. Assumption~\ref{ass:w_set} requires $\frac{\sqrt{\lambda^{(0)}} \max \limits_{s,i} y_{si}}{ \tau \lambda } \Vert \mathsf{W}^\top \mathsf{W} \Vert_2 < 1$, where $\sqrt{\lambda^{(0)}}$ is the standard deviation of the Gaussian likelihood. As $\tau \to 0$, this condition becomes impossible to satisfy for nontrivial $\mathsf{W}$. 

Instead of requiring that the inner constraint of a DEQ finds a unique global minimum, we relax the constraint so that it finds a local minimum. This allows us to use a weaker assumption that may be satisfied by the $\ReLU$.
\begin{assumption}
\label{ass:w_set_relu}
Suppose $R(\eta) = \ReLU(\eta)$. Let $a<\infty$ be a Lipschitz constant of $A'$. The support $\mathbb{W}$ satisfies
\begin{align*}
\mathbb{W} \subseteq \big\{ \mathsf{W} \in \mathbb{R}^{d \times l}, b \in \mathbb{R}^d \mid  \kappa \Vert \mathsf{W}^\top \mathsf{W} \Vert_2 < 1 \big\}, 
\end{align*}
where $\kappa = a \lambda^{-1}.$
\end{assumption}
An interesting property of the $\ReLU$ is that the dropout function (where it is defined) maps to a finite set of values. We may form and analyse a set of nicely-behaved fixed point equations for each dropout value instead of analysing one fixed point equation involving $\Theta$, whose derivative is poorly behaved. This allows us to avoid dealing with the non-continuous nature of the dropout function $\Theta$.
\begin{restatable}{theorem}{relushallow}
\label{thm:relu_shallow}
Suppose $R(\eta) = \ReLU(\eta)$ and fix some parameters $\big(\mathsf{W}, B \big)$ and data index $s$. Any stationary point $Z^\ast_s$ of objective~\eqref{eq:posterior_Z} is a solution to
\begin{align*}
     Z_s^\ast &= \frac{1}{\lambda} \mathsf{W}^\top \big( T(Y_s) \odot \Theta(\mathsf{W}Z_s^\ast + B) - \ReLU(\mathsf{W} Z_s^\ast + B) \big).
\end{align*}
Under Assumption~\ref{ass:w_set_relu}, there exists at least $1$ and at most $2^d$ stationary points, all of which are local minima.
\end{restatable}

The proof is given in Appendix~\ref{app:proofs}. As expected, the fixed point equations in Theorems~\ref{thm:general_shallow} and~\ref{thm:relu_shallow} are identical, up to a substitution and ignoring the nondifferentiable point. Dropout (where defined) maps to a finite set for other activations such as the Leaky $\ReLU$ and hard sigmoid, and Theorem~\ref{thm:relu_shallow} can likely be extended to these cases.
\subsection{Deep PED}
\label{sec:deep}
\begin{figure}
    \centering
        \begin{tikzpicture}[latent/.append style={minimum size=0.9cm},obs/.append style={minimum size=0.9cm}]
          % Define nodes
          \node[obs]                                (y) {$Y_s$};
          \node[latent, above=of y, xshift=0.5cm]   (w) {$\mathsf{W}^{(1)}$};
            \node[latent, above=of y, xshift=-0.5cm]   (b) {$B^{(1)}$};
          \node[latent, right=of y]                  (z) {$Z_s^{(1)}$};

        \node[latent, right=of z]                  (zdots) {$\ldots$};
        \node[latent, right=of zdots]                  (zlast) {$Z_s^{(L)}$};
          % Connect the nodes
          \edge {z,w,b} {y} ; %
          \edge {zdots} {z};
          \edge{zlast} {zdots};
          
        \node[latent, above=of z, xshift=0.5cm]   (w2) {$\mathsf{W}^{(2)}$};
        \node[latent, above=of z, xshift=-0.5cm]   (b2) {$B^{(2)}$};

        \node[latent, above=of zdots, xshift=0.5cm]   (weps) {$\ldots$};
        \node[latent, above=of zdots, xshift=-0.5cm]   (beps) {$\ldots$};
        
        \edge{b2,w2} {z};
        \edge{beps,weps} {zdots};
        
          % Plates
          \plate[inner sep=0.35cm] {yz} {(z)(y)(zdots)(zlast)} {$s=1,\ldots,N$} ;
        \end{tikzpicture} 
        \iffalse
            \begin{align*}
            Z_s^{(0)} := Y_s &\sim \mathcal{D}_{A, \lambda^{(0)}}\big(R^{(1)}(H_s^{(1)})\big), \quad H_s^{(1)} = \mathsf{W}^{(1)} Z_s^{(1)} + B^{(1)} \\
            Z_s^{(l)} &\sim \mathcal{D}_{A, \lambda^{(l)}}\big(R^{(l+1)}(H_s^{(l+1)})\big), \quad H_s^{(l+1)} = \mathsf{W}^{(l+1)} Z_s^{(l+1)} + B^{(l+1)}, \quad 1 \leq l < L \\
            Z_s^{(L)} &\sim \mathcal{N}\big(0, \mathsf{I}/\lambda^{(L)} \big) \qquad \mathsf{W}^{(l)}, B^{(l)} \sim \mathcal{P}^{(l)}( \mathbb{W}^{d^{(l)} \times l + d^{(l)}}), \quad 1 \leq l \leq L
        \end{align*}   \fi
    \caption{Restricting our attention to the Gaussian case $A(\eta) = \eta^2/2$, we consider a deep graphical model. Here each of the latent $Z^{(l)}$ variables are Gaussian conditioned on their parents, with variance $1/\lambda^{(l)}$ and canonical parameter $R^{(l+1)}\big( \mathsf{W}^{(l+1)}Z_s^{(l+1)} + B^{(l+1)} \big)$. Surprisingly, a DEQ can be used to find local minima in the latent variables.}
    \label{fig:bayes_nets_deep}
\end{figure}
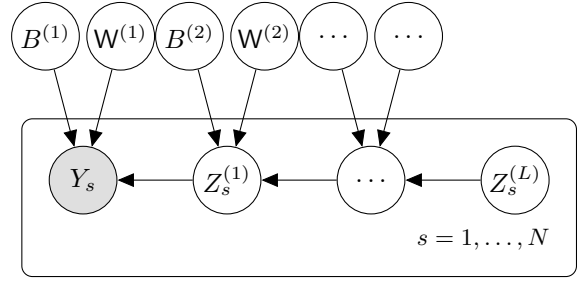
It is well-known that compositions of linear transformations are linear transformations. This means that stacking multiple PED layers when the likelihood is Gaussian and $R$ is the identity is trivial. When a nonlinearity is introduced, the analysis becomes challenging, but might offer the chance of more expressive neural network models due to the deep nonlinear structure of the architecture~\citep{poole2016exponential}.
As we now show, one may stack PED layers in a principled manner to obtain a deep PED architecture when $A$ is quadratic and $R$ is nonlinear. 

We consider a natural generalisation of Figure~\ref{fig:bayes_nets} in Figure~\ref{fig:bayes_nets_deep}. Instead of a single latent prior node, we have $L$ latent nodes built as a hierarchical prior. Here every object is given a superscript denoting the layer $1 \leq l \leq L$ to which it belongs. The conditional distribution of the latent node in layer $l$ is Gaussian in exponential family form $\expfam$, having $A(r) = r^2/2$ and an additional standard deviation parameter $\nu = \sqrt{\lambda^{(l)}}^{-1}$. Variables carrying over from the shallow case in \S~\ref{sec:shallow} are renamed $Y_s = Z_s^{(0)}$, $\lambda = \lambda^{(1)}$ and $(\mathsf{W}, B) = (\mathsf{W}^{(1)}, B^{(1)})$. Let $\theta = \{ \mathsf{W}^{(l)}, B^{(l)}\}_{l=1}^L$. With $s$ fixed, the joint posterior over $\theta, \{Z^{(l)}\}_{l=1}^L$ of the graphical model in Figure~\ref{fig:bayes_nets_deep} satisfies
\begin{align*}
    &\phantom{{}={}}p\big( Z^{(1)}, \ldots, Z^{(L)}, \theta  \mid Y \big) \label{eq:deep_posterior} \numberthis \\
    &\propto p\big(Y \mid Z^{(1)}, \theta \big)  \prod_{l=1}^{(L-1)} p\big(Z^{(l)} \mid Z^{(l+1)}, \theta\big) p(Z^{(L)}) p(\theta).
\end{align*}
As in the shallow case, we find the fixed point condition implied by the stationary points of the posterior, leading to a DEQ layer. Let $\zeta = (Z^{(1)}; \ldots; Z^{(L)} ) \in \mathbb{R}^D$ be an augmented state and write 
\begin{align*}
    &\phantom{{}={}} G^{[1:L]}(\zeta) \\
    &= \Big( G^{(1)}\big( Z^{(1)}; Z^{(0)}, Z^{(2)} \big); \ldots ;G^{(L)}\big( Z^{(L)}; Z^{(L-1)}, 0 \big)  \Big) \label{eq:deepPED} \numberthis
\end{align*}
where
\begin{align*}
    &\phantom{{}={}} G^{(l)}\big( Z^{(l)}; Z^{(l-1)}, Z^{(l+1)} \big) \\
    &= \frac{1}{\sqrt{\lambda^{(l)}}} R^{(l+1)}\big(W^{(l+1)} {Z^{(l+1)}} + b^{(l+1)}\big) + \\
    & \quad \frac{1}{\lambda^{(l)}}   {W^{(l)}}^\top  \Bigg( \rho^{(l)} ( W^{(l)}{Z^{(l)}} +b^{(l)})  \odot   \sqrt{\lambda^{(l-1)}}  {Z^{(l-1)}}   \\
    &\quad -  \sigma^{(l)}\big(   W^{(l)}{Z^{(l)}}+b^{(l)}  \big) \Bigg).
\end{align*}
In Appendix~\ref{app:deep}, we show that fixed points of~\eqref{eq:deepPED} are the stationary points of~\eqref{eq:deep_posterior} under certain conditions.
This augmentation of the state space shares an interesting connection to the construction used to argue that a single layer DEQ is sufficient to represent multiple DEQ layers~\citep[Theorem 2]{bai2019deep}.
Interestingly, due to the way we constructed the augmented state, we can identify different subsets of coordinates as corresponding with latent variables at different layers $l$. \emph{To the best of our knowledge, this is the first such DEQ with this interpretable quality.}
Unfortunately, formally describing the nature of the stationary points which are the fixed points of $G^{[1:L]}$ is difficult --- a limitation of our work. 
See Appendix~\ref{app:deep} for a sketch. 

\iffalse
Solving this fixed point equation in $Z^{(l)}$ requires knowing $Z^{(l-1)}$ and $Z^{(l+1)}$. Unfortunately, we solve each fixed point equation in ascending $l$, we know $Z^{(l-1)}$ but not $Z^{(l+1)}$. We may find a local MAP over $Z^{(1)}, \ldots, Z^{(L)}$ by applying the iterated conditional modes algorithm initialised at the zero vector, as shown in \texttt{PED.forward()}, Algorithm~\ref{alg:map}. 
The parameters can be jointly adjusted together with the latents found by wrapping \texttt{PED.forward()} in a parameter adjustment scheme that attempts to solve the problem
\begin{equation}
\begin{aligned}
    \min_{\theta} \quad &\sum_{s=1}^T \frac{\gamma}{2} \Vert \theta \Vert^2 - T(Y_s)^\top \tReLU(\eta_s) + \frac{1}{2}1^\top \tReLU^2(\eta_s)  \\
    \text{subject to } \quad & \eta_s = W^{(1)} Z_s^{(1)} + b^{(1)} \\
    &\{ {Z_s^{(l)}}^\ast \}_{l=1}^L = \texttt{PED.forward}(Y_s, \theta, C), \qquad s = 1, \ldots, N.
\end{aligned}
\label{eq:ped_learning}
\end{equation}
It is interesting to consider the special case $C=2$, where the iteration $c=1$ can be interpreted as a pass through an encoder, and the iteration $c=2$ can be interpreted as a pass through a decoder. The outer optimisation problem in~\eqref{eq:ped_learning} then represents a global regularised reconstruction loss.

\subfile{../figures/algorithm}
\fi

\subsection{Related work and other considerations}
\label{sec:related-work}
Several theoretical issues of DEQs have been investigated. These include methods for ensuring certified rosbustness~\citep{wei2021certified}, constraints and architectures that ensure or help well-posedness of the fixed point condition~\citep{winston2020monotone,revay2020lipschitz,el2021implicit} and connections to nonparametric statistical estimators~\citep{tsuchida2021declarative}.~\citet{li2021optimization} design DEQ layers from a (non-statistically motivated) optimisation perspective. We note that Jacobian regularisation~\citep{bai2021stabilizing} may be used as a soft penality on the weights to allow Assumptions~\ref{ass:w_set},~\ref{ass:w_set_relu} hold, where required. Implicit networks offer the opportunity to inject a degree of interpretability into neural network models, if the problem that the implicit layer solves is itself interpretable.
This is especially true for optimisation-based layers~\citep{monga2021algorithm}, but not necessarily for DEQ layers.
Our work identifies problems which DEQ layers solve, and shows that such problems have meaningful statistical interpretations.

\paragraph{Linearly parameterised exponential families.} Exponential families are widely used due to their flexibility and the strict convexity and smoothness properties of the log likeihood of~\eqref{eq:exp_fam} in $r$. Typically, the canonical parameters are set to be a parameterised linear transformation of an input (in supervised settings~\citep{mccullagh1989generalized,loader2006local}) or a latent variable (in unsupervised settings~\citep{collins2001generalization,mohamed2008bayesian}). This extends when the linear transformation represents an element of an RKHS~\citep{canu2006kernel}. Linear transformations compose well with exponential families because they preserve convexity (in the parameters) and smoothness.

\begin{figure}
    \centering
    \includegraphics[scale=0.25]{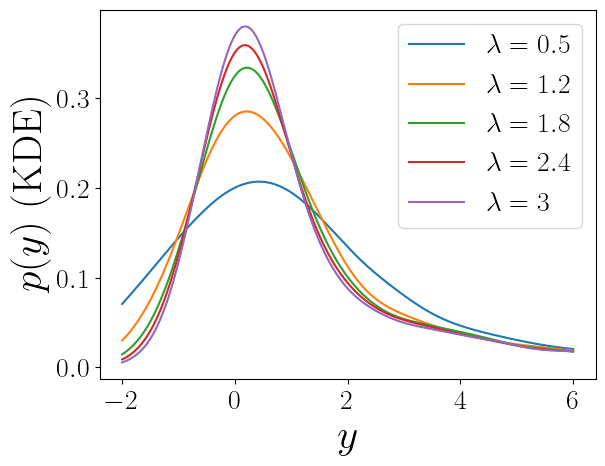}
    \caption{KDE of~\eqref{eq:relu_gaussian} in $1$ dimension with $\tau = 0.5$ using $10000$ samples of $\mathsf{W}$, $z$ and $y$.}
\end{figure}

\paragraph{$A$ is not all you need.} Using a non-identity $R$ allows for a number of potential benefits. If $R$ is chosen to be the identity, the activation function $\sigma$ must be increasing by strict convexity of $A$. Therefore, activations such as the $\ReLU$, GELU, nonconstant periodic functions, or Gaussian activations are not obtainable with linear $R$, each of which have been shown to posses useful properties~\citep{nair2010rectified,hendrycks2016gaussian,NEURIPS2021_0d5a4a5a} (although perhaps in slightly different architectures). We further detail this lack of expressiveness for $A$ for the softplus activation in Appendices~\ref{app:einstein1} and~\ref{app:einstein2}. 

Another advantage of nonlinear $R$ is that it can map inputs into an acceptable range for the canonical parameter of the exponential family. For example, an exponential distribution requires a negative canonical parameter, which can be ensured by choosing $R(\eta) = - \tReLU(\eta)$, which may be easier than constraining $\eta$ to be negative. 

Finally, $R$ allows us to model situations that cannot be modelled using the linearly parameterised exponential family.
For example, let $A(r) = r^2/2$ and $R(\eta) = \tReLU(\eta)$. In order to visualise what this means, consider the case $d=l=1$. With a $\mathcal{N}(0, \sqrt{2})$ prior over $\mathsf{W} \in \mathbb{R}$ and $z \in \mathbb{R}$,
\begin{align*}
    p(y) &= \mathbb{E}_{\mathsf{W}, z \sim \mathcal{N}(0,1)} \big[p\big(y \mid \tReLU( \mathsf{W} z) \big)   \big] \numberthis \label{eq:relu_gaussian}
\end{align*}
is visualised by sampling $10000$ $\mathsf{W}$, $z$ and $y$ and plotting a KDE for $y$. This is shown in Figure~\ref{fig:gauss_kde}, where the variance of the conditional distribution of $y$ given $\mathsf{W}z$ is $\lambda^{-1}$.

\paragraph{A recipe for mapping $\sigma$ and $\rho$ to nonlinearly parameterised Gaussians.} Using the various relations between $A$, $R$, $\sigma$ and $\rho$, it is possible to understand soft dropout and activation functions in terms of nonlinearly parameterised exponential family likelihoods. We demonstrate this for quadratic $A(v)=v^2/2$ and nonlinear $R$. We have by definition that $\sigma(\eta) = (A \circ R)'(\eta) = \frac{1}{2} \frac{d}{d\eta} R^2(\eta)$, so that 
\begin{equation}
    \big| R(\eta) \big| = \sqrt{\int 2 \sigma(\eta) \, d\eta }, \quad \rho(\eta) = \sigma(\eta)/R(\eta). \label{eq:recipe}
\end{equation}

As discussed above and in Appendices~\ref{app:einstein1} and~\ref{app:einstein2}, if $\sigma$ is the commonly used softplus activation, it does not have an interpretation as the derivative of a log-partition function. However, we may understand it through a nonlinear transformation $R(\eta) = \sqrt{\int 2 \log\big(1+\exp(\eta)\big)  \, d\eta } = \sqrt{-2 \text{Li}_2(-e^\eta) }$ of the canonical parameter of a Gaussian distribution, allowing it to model real-valued data. Here $\text{Li}_2$ denotes the dilogarithm function. See Appendix~\ref{app:einstein2} for details. See Figure~\ref{fig:r_plots} for plots of $R$ and $\rho$.

This construction can also be used to understand instances of $\sigma$ that \emph{are} derivatives log-partition functions of non-Gaussian exponential families alternatively as arising from nonlinear transformations of Gaussians. For example, as is well-known, the logistic sigmoid is the inverse-link function of a Bernoulli random variable, suggesting that it should be used to model binary-valued data. However, by~\eqref{eq:recipe}, we may understand it through a nonlinear transformation $R(z) = \sqrt{\int 2 \big(1+\exp(-z)\big)^{-1}  \, dz } = \sqrt{2 \log(1+e^z)}$ of the canonical parameter of a Gaussian distribution, allowing it to model real-valued data. See Appendix~\ref{app:sigmoid} for details.  See Figure~\ref{fig:r_plots} for plot of $R$ and $\rho$.

\vspace{-0.5cm}
\paragraph{Bregman divergences.} The negative log-likelihood of an exponential family can always be written as a Bregman divergence $B_{\phi}(y, \mu)$ in terms of its expectation parameters $\mu$ (with an additional factor that is constant with respect to the expectation parameter) and convex conjugate $\phi$ of $A$~\citep[Theorem 4]{banerjee2005clustering}. 
For each exponential family, there is a unique Bregman divergence.
For example, the Gaussian corresponds with the squared loss, the Poisson corresponds with relative entropy, and Exponential  corresponds with the Itakura Saito divergence. 
Under our nonlinear parameterisation, the expectation parameter is $\mu = A' \circ R \, (\eta)$, leading to a space of nonlinearly parameterised Bregman divergences. 
See Appendix~\ref{app:bregman}.

\begin{figure}
    \centering
    \includegraphics[scale=0.25]{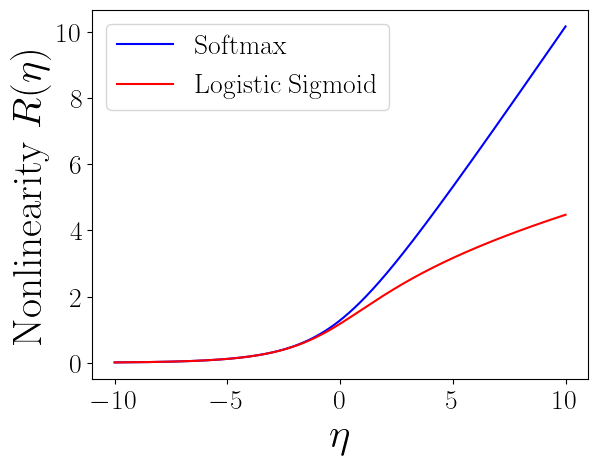}
    \includegraphics[scale=0.25]{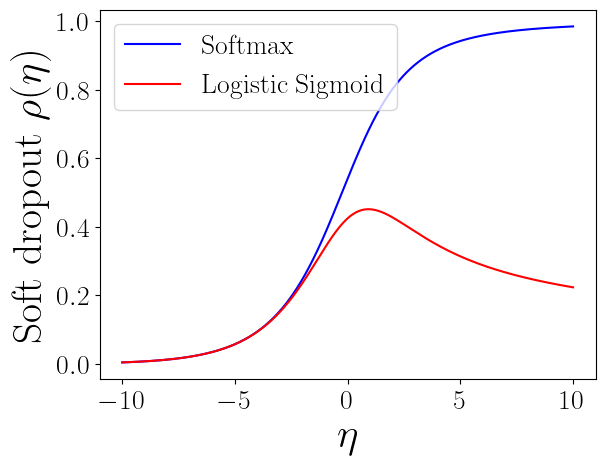}
    \caption{Choosing $\sigma$ to be logistic sigmoid or softmax, the nonlinearity $R$ and soft dropout function $\rho$, in~\eqref{eq:recipe}.}
    \label{fig:r_plots}
\end{figure}

\iffalse Let $\mathbb{S}_{\succeq 0}(l) = \{ P \in \mathbb{R}^{l \times l} \mid P = P^\top, v^\top P v \geq 0\,  \forall v \in \mathbb{R}^l \}$ denote the space of $l \times l$ symmetric positive semidefinite (PSD) matrices. Correspondingly let $\mathbb{S}_{\succ 0}$ denote the space of strictly PD matrices. Let $\mathbb{O}(l) = \{ O \in \mathbb{R}^{l \times l} \mid O^\top O = O O^\top = I  \}$ denote the set of orthogonal matrices, and let $\mathbb{O}(d, l) = \{ O \in \mathbb{R}^{d \times l} \mid O^\top O = I  \}$ denote the set of ``semi-orthogonal" matrices. Every $O \in \mathbb{O}(l)$ forms an orthonormal basis for $\mathbb{R}^l$ and every $O \in \mathbb{O}(d, l)$ forms an orthnormal basis for an $l$ dimensional linear subspace of $\mathbb{R}^d$.\fi

\iffalse
\begin{lemma}
\label{lem:closure}
Given any fixed $Q \in \mathbb{O}(l)$, $R \in O(d, l)$ if and only if there exists a $B \in \mathbb{O}(d,l)$ such that $R = BQ$.
\end{lemma}
\begin{proof}
If $R=BQ$ for some $B \in \mathbb{O}(d,l)$, then $R^\top R = B^\top Q^\top Q B^\top = I$, so $R \in \mathbb{O}(d,l)$. 

Conversely if $R \in (d,l)$, then $\min_{B \in \mathbb{O}(d,l)} \Vert BQ - R \Vert_2 = \min_{B \in \mathbb{O}(d,l)} \Vert B - \underbrace{RQ^\top}_{\in \mathbb{O}(d,l)} \Vert_2 = 0$, so there exists a $B \in \mathbb{O}(d,l)$ such that $R=BQ$.
\end{proof}
\fi

\textbf{Convolutional layers} may be represented as sparse matrices with repeated entries. Our results apply when $\mathbb{W}$ is a space of convolutional layers. The transpose operator can be computed using transposed convolution~\citep{zeiler2010deconvolutional}. The constraints in Assumption~\ref{ass:w_set} and~\ref{ass:w_set_relu} may be computed efficiently when the norm is the spectral norm~\citep{sedghi2018the}. See Appendix~\ref{app:conv}.

\section{Illustration on 2D latent space}
\begin{wrapfigure}[9]{l}{0.14\textwidth}
    \centering
    \includegraphics[scale=0.15]{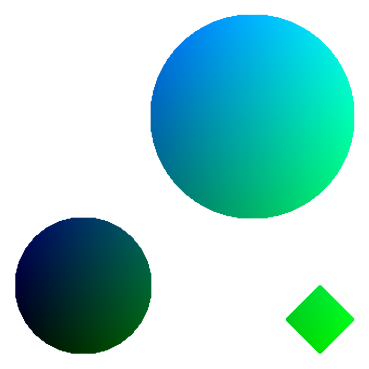}
    \caption{\\ground truth, $\mathsf{Z}$}
    \label{fig:true_shape}
\end{wrapfigure} 
In this section, we compare the performance and key features of PCA, tSNE~\citep{van2008visualizing}, UMAP~\citep{mcinnes2018umap} and PED.
It can be difficult to evaluate latent projections. 
Our evaluation focuses on two issues:
visualising the latent space, and performance in a downstream classifier.

\begin{figure*}[!ht]
    \centering
    \begin{tabular}{b{0.3\textwidth} | c c c c}
         & PCA (sklearn) & tSNE & UMAP & PED  \\ \hline
         \begin{tabular}[b]{l}
         Gaussian\\ $A(r)= r^2/2$ \\
         $R(\eta) = \eta$\end{tabular}& 
         \includegraphics[scale=0.15]{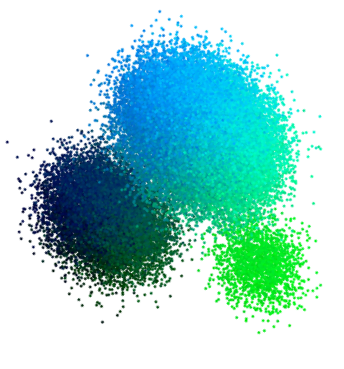} 5 &  
         \includegraphics[scale=0.15]{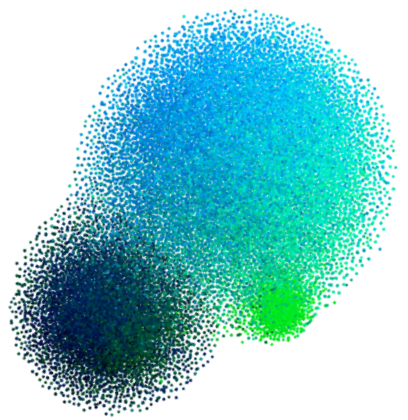} - & 
         \includegraphics[scale=0.15]{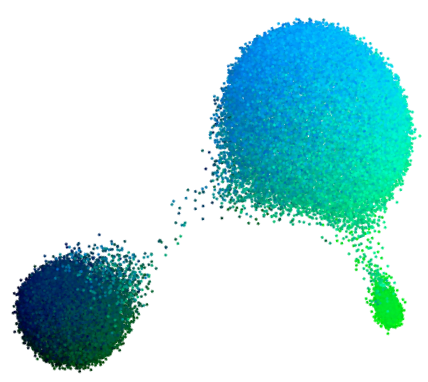} 0 & 
         \includegraphics[scale=0.15]{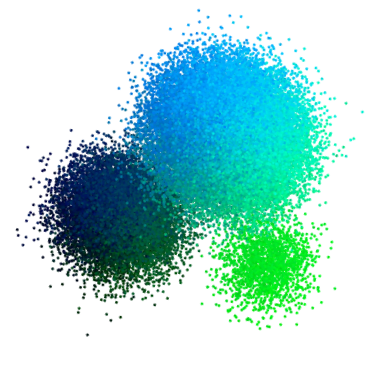} \textbf{95 }\\ \hline 
         \begin{tabular}[b]{l}
         Bernoulli\\ $A(r)= \log \big(1+e^r \big)$ \\
         $R(\eta) = \eta$\end{tabular} & 
         \includegraphics[scale=0.15]{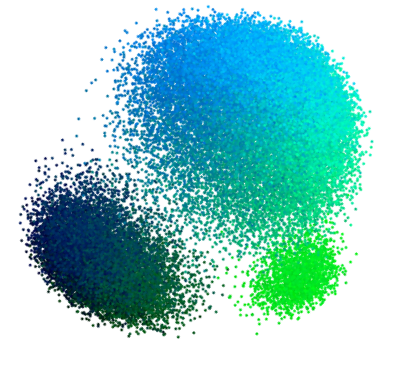} 6 &  
         \includegraphics[scale=0.15]{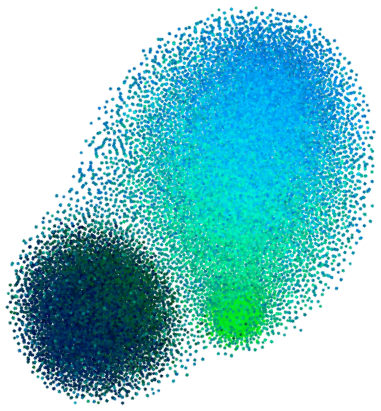} - & 
         \includegraphics[scale=0.15]{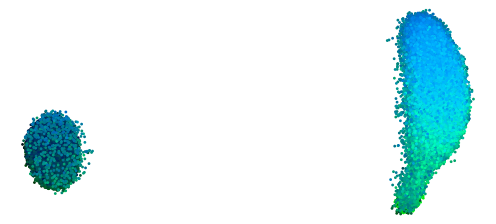} 0 & 
         \includegraphics[scale=0.15]{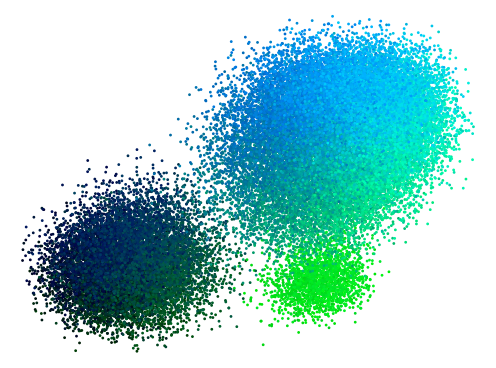} \textbf{94 }\\ \hline 
         \begin{tabular}[b]{l}
         Poisson\\ $A(r)= \exp(r)$ \\
         $R(\eta) = \eta$\end{tabular} & 
         \includegraphics[scale=0.15]{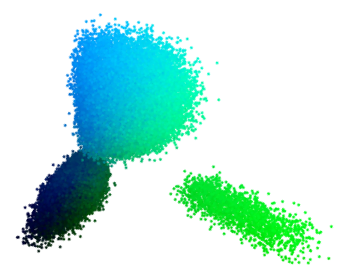} 0 &  
         \includegraphics[scale=0.15]{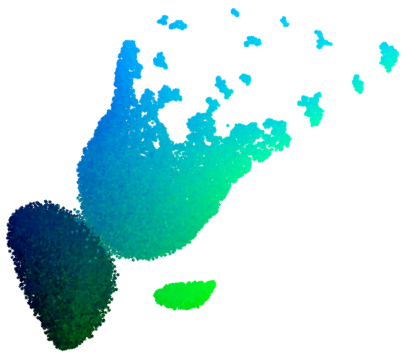} - & 
         \includegraphics[scale=0.15]{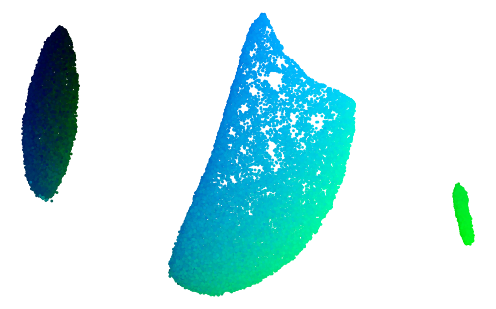} 16 & 
         \includegraphics[scale=0.15]{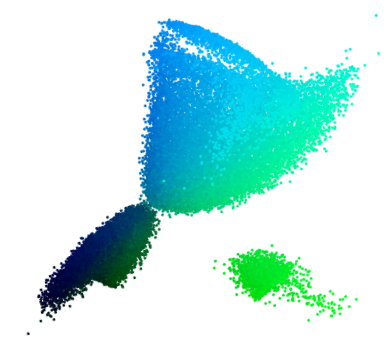} \textbf{84 }\\ \hline 
         \begin{tabular}[b]{l}
         Rectified-mean Gaussian (shallow)\\ $A(r)= r^2/2$ \\
         $R(\eta) = \ReLU(\eta)$ \\
         $L = 1$\end{tabular} & 
         \includegraphics[scale=0.15]{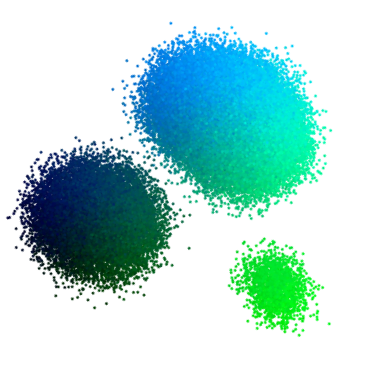} 0 &  
         \includegraphics[scale=0.15]{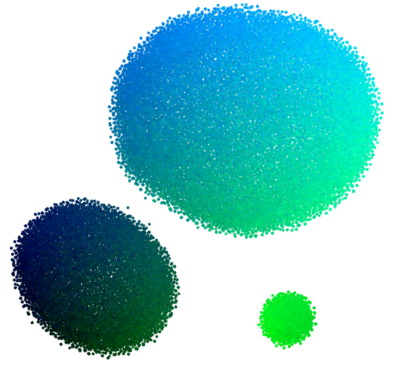} - & 
         \includegraphics[scale=0.15]{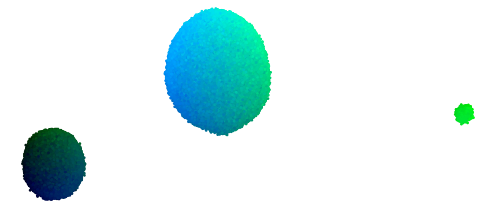} 11 & 
         \includegraphics[scale=0.15]{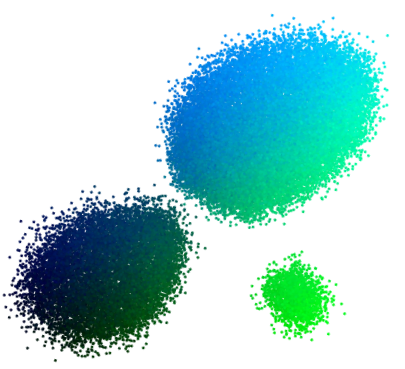} \textbf{89 }\\ \hline 
         \begin{tabular}[b]{l}
         Rectified-mean Gaussian (deep)\\ $A(r)= r^2/2$ \\
         $R(\eta) = \ReLU(\eta)$ \\
         $L = 2,\quad
         d^{(1)} = 30, d^{(2)} = 2$\end{tabular} & 
         \includegraphics[scale=0.15]{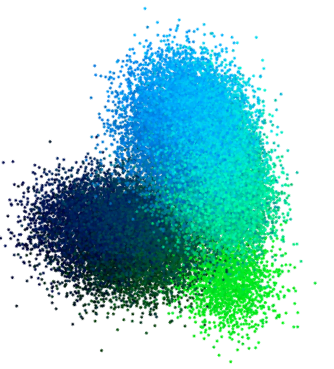} 0 &  
         \includegraphics[scale=0.15]{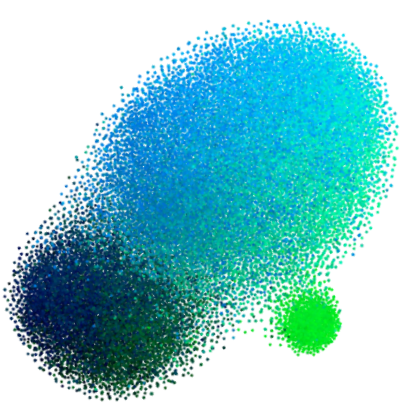} - & 
         \includegraphics[scale=0.15]{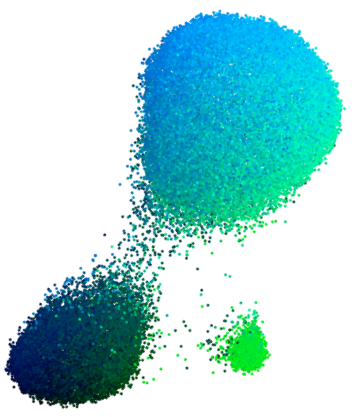} 0 & 
         \includegraphics[scale=0.15]{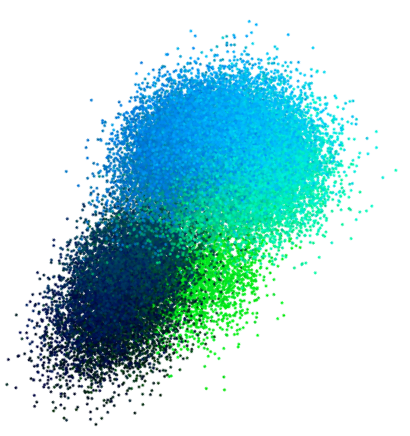} \textbf{100} \\ \hline 
    \end{tabular}
    \caption{\looseness=-1  The left column describes the ground truth distribution, with latents as shown in Figure~\ref{fig:true_shape}. The number in each cell represents the number of times that model performed the best compared with others when used as the input for a downstream task over $100$ random trials. Interestingly, PCA and PED often produce visually quite similar results to each other and the ground truth. However, PED is able to obtain much better results on downstream tasks on account of its ability to be fine-tuned. While UMAP's results are visually pleasing, its embeddings do not bear much similarity with the ground truth, except in the Gaussian case. For non-Gaussian cases, UMAP sometimes places latents in good clusters, but fails to globally position the clusters accurately relative to one another. tSNE is incompatible in a pipeline with a downstream task. tSNE produces globally accurate positioning, but sometimes adds artefacts into the visualisation. Embeddings are an equivalence class up to a rotation and scaling, since $\mathsf{W}$ and $\mathsf{Z}$ are non-identifiable; we rotated images by hand to align them. For PED, we visualise the latents in an orthogonal basis, i.e. shown are $\mathsf{R} \mathsf{Z}$, where $\mathsf{W} = \mathsf{Q} \mathsf{R}$ is a QR decomposition.}
    \label{fig:synthetic}
\end{figure*}

 \paragraph{Ground truth latents.} We generate datasets through graphical model in Figure~\ref{fig:bayes_nets}, except that $\mathsf{Z}$ is fixed to be as shown in Figure~\ref{fig:true_shape}. 
For each distribution shown in the left column of Figure~\ref{fig:synthetic}, we generate $100$ parameters $\mathsf{W}$ according to the graphical model and them sample the corresponding $\mathsf{Y}$.
This results in $100$ different datasets for each distribution, each with known ground truth $\mathsf{Z}$.

\paragraph{Downstream performance and deep learning compatibility.} A key feature of PED is that it is compatible with any other neural network layer.
This means that we can fine-tune backbone PED layers that are pretrained in an unsupervised manner on a supervised task with a head network.
In contrast, since PCA is a linear mapping, composing it with a head network and optimising the result is equivalent to just using the head network.
Similarly, other dimensionality reduction techniques that do not posses neural network parameters such as UMAP cannot benefit from fine-tuning. tSNE, which is better thought of as a visualisation tool than a dimensionality reduction tool, is even less suitable in this respect. The mapping produced by tSNE on training data cannot be used on testing data, so it is incompatible in a pipeline with downstream tasks.

We use a simple fully connected head with widths $2, 100, 1$ and $\ReLU$ activations at width $100$. Together with a PED or (untrainable) PCA/UMAP backbone, we train the full network to predict $g(Z_s) \in \mathbb{R}$ given an input $Y_s \in \mathbb{R}^{50}$ by minimising the sum of squared errors. This is done for each of the $100$ randomly generated datasets. The corresponding $Z_s \in \mathbb{R}^2$ are not known to the network. This is a simple to visualise and deceptively difficult task that relies heavily on the quality of the latent embedding. Full hyperparameter details and further experiments are given in Appendix~\ref{app:experiments}.

A limitation of our evaluation is that we did not tune hyperparameters of any algorithm. However, if latents were to be used in a downstream task that was not known \textit{a priori}, there would be no way of objectively tuning hyperparameters. We have demonstrated the important benefit that PED has over all other techniques --- it is end-to-end differentiable and can be fine-tuned. PED  motivates use of DEQ layers for supervised tasks as using fine-tuned latents.
\section{Discussion and conclusion}
\label{sec:conclusion}
We proposed a canonical nonlinear map for parameters of an exponential family, and considered the problem of MAP estimation in PCA.
We derived shallow and deep DEQ architectures that solve this problem, and called these shallow and deep PED respectively.
Our main contribution is the theoretical analysis. 
It grounds DEQ architectures in a probabilistic framework, and shows how certain architectural choices can be related back to statistical assumptions on the observations.
More generally, use of DEQ layers admitting the same form as PED layers can be viewed as fine-tuning for a supervised task.

As noted by~\citet{bai2019deep}, DEQs built from contraction mappings may be understood as infinitely deep neural networks with shared weights.
In this sense, our analysis motivates the use of such infinitely deep networks.
Machine learning has a history of taking algorithms that at first seem to require infinitely many samples or steps and running those algorithms for one or very few steps~\citep{hinton2002training,sutskever2010convergence,kingma2013auto}.
In future, it might be possible that few steps of a DEQ forward iteration can be understood using the theory of infinitely many forward iterations, just as well-behaved wide neural networks can now be understood using the theory of infinitely wide neural networks~\citep{jacot2018neural}.

Another direction is to extend PED to a Bayesian setting by taking the Hessian of the negative log likelihood evaluated at the MAP.
Expressions of second order approximations are isolated in Appendix~\ref{app:bayesian}, and may be useful for future investigations, for example in Laplace approximation~\citep{daxberger2021laplace}.
Interestingly, in some cases the curvature at the MAP depends on a deterministic dropout, which offer another~\citep{srivastava2014dropout, pmlr-v48-gal16} perspective on dropout.

%Empirical success stories of DEQs applied to language prediction and computer vision usually other deep learning elements such as activations that are not canonical inverse link functions (such as the ReLU), attention and/or multi-scale residual blocks~\citep{bai2020multiscale}. In future, we hope to cast these elements in light of MAP estimation.

%Although we implicitly focused on the setting where $l < d$ in the shallow case, and $d^{(0)}<d^{(1)}\ldots<d^{(L)}$ in the deep case, none of our analysis actually requires this. This allows one to construct general DEQ layers for use in unsupervised tasks, which may be interpreted as fine-tuned nonlinearly parameterised exponential family PCA.

We hope that our link between neural network design choices (activation functions, dropout, and layer structure) and their corresponding statistical distribution counterparts (log partition function, canonical nonlinearity) enables other researchers to gain a better understanding of equilibrium networks.

\bibliography{pca}

\newpage
\clearpage
\onecolumn
\appendix

\section{The special case of Gaussian distributions}
\label{app:gaussian}
In this section, we work under the assumption is that the exponential family model is a Gaussian distribution in order to illustrate a simple special case of Theorem~\ref{thm:general_shallow}. We follow~\cite[\S 10.7]{deisenroth20mathml} for probabilistic PCA, with some small differences. Firstly, our setup allows for a prior over joint $\mathsf{W}, B$, and we consider point estimates of $Z_s$. Secondly, our prior over $\mathsf{Z}$ has variance $\lambda^{-1}$ and the Gaussian likelihood has unit variance, whereas in ~\cite[\S 10.7]{deisenroth20mathml} $\mathsf{Z}$ has a unit variance and the Gaussian likelihood has variance $\sigma^2$. As we shall see, our resulting MAP estimates are the same, up to this scaling reparameterisation.

For the Gaussian, we have $T(y) = y/b$, where $b>0$ is a fixed scale parameter, $A(v) = v^2/2$, and $R=1$. Without loss of generality, we take $b=1$, since input data $\mathsf{Y}$ may be preprocessed to account for different scaling. Let $\Vert \cdot \Vert_F$ denote the Frobenius matrix norm. Following~\eqref{eq:nonlinear-expfam}, the negative log posterior of~\eqref{eq:posterior_Z} is
\begin{align*}
    - \log p(\mathsf{Z} \mid \mathsf{W}, B, \mathsf{Y}) &= - \log h(\mathsf{Y}) + \frac{\lambda}{2} \Vert \mathsf{Z} \Vert^2_F + \sum_{s=1}^N \sum_{i=1}^d \frac{1}{2} (\mathsf{W} Z_s + B)_i^2- y_{si} (\mathsf{W} Z_s + B)_i.
\end{align*}
This is a strongly convex objective, whose derivative is zero at the unique minima $\mathsf{Z}^\ast$. Differentiating with repect to $Z_s$, we have that the minimiser $\mathsf{Z}^\ast$ satisfies
\begin{align*}
    0 &= \lambda Z_s^\ast + \mathsf{W}^\top \Big( \mathsf{W} Z_s^\ast + B - Y_s  \Big) \\
    Z_s^\ast &= \frac{1}{\lambda} \mathsf{W}^\top \Big( Y_s -  (\mathsf{W} Z_s^\ast + B) \Big). \numberthis \label{eq:gaussian_constraint}
\end{align*}
This is a DEQ with linear activation having evaluation $(\mathsf{W} Z_s + B)$. The dropout function is $1$. The reprojection error loss is obtained by considering the minimiser of the joint posterior
\begin{align*}
    - \log p(\mathsf{Z} , \mathsf{W}, B \mid \mathsf{Y}) &= -\log p(\mathsf{W}, B) - \log p(\mathsf{Z} \mid \mathsf{W}, B, \mathsf{Y}) \\
    \mathsf{Z}^\ast, \mathsf{W}^\ast, B^\ast &= \argmin_{\mathsf{Z}, \mathsf{W}, B} -\log p(\mathsf{W}, B) - \log p(\mathsf{Z} \mid \mathsf{W}, B, \mathsf{Y}) \\
    \mathsf{W}^\ast, B^\ast &= \argmin_{\mathsf{W}, B} -\log p(\mathsf{W}, B) - \log p(\mathsf{Z}^\ast \mid \mathsf{W}, B, \mathsf{Y}), \quad \text{subject to} \quad \mathsf{Z}^\ast = \argmin_{\mathsf{Z}} - \log p(\mathsf{Z} \mid \mathsf{W}, B, \mathsf{Y}).
\end{align*}
The constraint is equivalent to~\eqref{eq:gaussian_constraint}. Substituting for the outer minimisaton problem, we have
\begin{align*}
    \mathsf{W}^\ast, B^\ast &= \argmin_{\mathsf{W}, B} -\log p(\mathsf{W}, B) + \sum_{s=1}^N \sum_{i=1}^d \frac{1}{2} (\mathsf{W} Z_s + B)_i^2- y_{si} (\mathsf{W} Z_s + B)_i \\
    \text{subject to} \quad Z_s^\ast &= \frac{1}{\lambda} \mathsf{W}^\top \Big( Y_s -  (\mathsf{W} Z_s^\ast + B) \Big) \quad \forall s=1, \ldots, N.
    \intertext{Or, more explicitly as a squared reprojection error,}
        \mathsf{W}^\ast, B^\ast &= \argmin_{\mathsf{W}, B} -\log p(\mathsf{W}, B) + \frac{1}{2} \sum_{s=1}^N \sum_{i=1}^d  \Big( (\mathsf{W} Z_s + B)_i - y_{si} \Big)^2 \numberthis \label{eq:reprojection_gaussian} \\ 
    \text{subject to} \quad Z_s^\ast &= \frac{1}{\lambda} \mathsf{W}^\top \Big( Y_s -  (\mathsf{W} Z_s^\ast + B) \Big) \quad \forall s=1, \ldots, N.
\end{align*}
In this special case, we may rearrange the constraint to a more familiar form with the help of the push-through (Woodbury) matrix identity,
\begin{align*}
    \lambda Z_s^\ast + \mathsf{W}^\top\mathsf{W} Z_s^\ast &= \mathsf{W}^\top (Y_s - B) \\
    Z_s^\ast &= \big( \lambda \mathsf{I} + \mathsf{W}^\top \mathsf{W} \big)^{-1} \mathsf{W}^\top (Y_s - B) \\
    Z_s^\ast &= \mathsf{W}^\top \big( \lambda \mathsf{I} + \mathsf{W} \mathsf{W}^\top \big)^{-1} (Y_s - B),
\end{align*}
which is the same as Equation (10.74) of~\cite{deisenroth20mathml}, up to a scaling reparameterisation.

Finally, we note a difference in approaches typically identified as ``latent variable models" and ``neural network autoencoder models". In the former, the marginal distribution $p(\mathsf{Y} \mid \mathsf{W}, B)$ or $p(\mathsf{W}, B \mid \mathsf{Y})$ obtained by integrating out latent variables $\mathsf{Z}$ is used as an objective for point estimation of parameters $\mathsf{W}$ and $B$ (see \citet[Remark p. 342]{deisenroth20mathml}). In the latter, a reprojection error like~\eqref{eq:reprojection_gaussian}, which depends on the latent variable, is used as an objective for estimation the parameters $\mathsf{W}$ and $B$. 

\clearpage
\section{Exponential family calculations}
\label{app:log_partition}

\subsection{Construction of exponential families}
\label{sec:expfam-construction}

Let $\mathbb{Y} \subseteq \mathbb{R}$ and form a measurable space $(\mathbb{Y}, \mathcal{B})$ for some sigma algebra $\mathcal{B}$. Let $\omega$ be some reference measure (nominally Lebesgue or counting) and let $h:\mathbb{Y} \to \mathbb{R}_{\geq 0}$ be some $\omega$-integrable function, $\int_{\mathbb{Y}} h(y) d\omega(y) < \infty$.  Let $T$ be some measurable function and let $\mathbb{F}\subseteq \mathbb{R}$ be the set of all canonical parameters ${r}$ such that $\int_{\mathbb{Y}} \exp\big( T(y) {r}) h(y) d\omega(y) < \infty$. Assume that $\mathbb{F}$ is open. Define the \emph{log partition function} $A:\mathbb{F}\to\mathbb{R}$ by $A({r}) = \log \int_{\mathbb{Y}} \exp\big( T(y) {r}) h(y) d\omega(y)$. Call 
\begin{align*}
    p(y \mid {r}) = h(y) \exp\big( {r} T(y) - A({r}) \big) \numberthis \label{appeq:exp_fam}
\end{align*}
the PDF of a minimal regular exponential family. For such a family, $A$ is both infinitely differentiable and strictly convex~\citep[Proposition 3.1]{wainwright2008}. Strictly convex and infinitely differentiable functions $A$ are admissible if and only if ${\exp\big( A(i {r}) \big)}$ is a positive definite function. See Lemma~\ref{lem:admissible}, Appendix~\ref{app:admissible}.
$A$ acts as a cumulant generating function, so that in particular the expected value under $p(y \mid r)$ is $\mu = A'(r)$. The parameter $\mu$ is called the expectation parameter.
Following common convention, we will use $p( \cdot \mid \cdot )$ to denote all (conditional) probability density functions, with a meaning clear from the arguments. We write $y \sim \mathcal{D}_{A,\chi}(r)$ to mean that a scalar-valued random variable $y$ follows an exponential family with log partition function $A$, scalar canonical parameter $r$ and additional parameter $\nu$\footnote{Some exponential families have an additional parameter that is not a canonical parameter. For example, Gaussian has a scale parameter and Laplace has a centering parameter. Both $A$ and $\nu$ are fixed and not learned.}. If $Y=(y_1, \ldots, y_d)$ and $V = (r_1, \ldots, r_d)$, we write $Y \sim \mathcal{D}_{A,\chi}(V)$ to mean $y_i \sim \mathcal{D}_{A,\chi}(r_i)$ for all $1 \leq i \leq d$, with each $y_i$ mutually conditionally independent given $H$.

\subsection{How to check whether a log partition function is admissible}
\label{app:admissible}
\begin{restatable}{lemma}{admissible}
An infinitely differentiable and strictly convex $A$ is an admissible log partition function for a minimal, regular, one-dimensional exponential family if and only if 
\begin{itemize}
\item there exists some $c$ such that $c \exp\big( A({r}) \big)$ is the moment generating function of some random variable evaluated at ${r} \in \mathbb{F} \subseteq \mathbb{R}$.
\item ${\exp\big( A(i {r}) \big)}$ is a positive definite function evaluated at ${r} \in \mathbb{R}$ and is continuous at ${r}=0$.
\end{itemize}
\label{lem:admissible}
\end{restatable}
\begin{proof}
Define
\begin{align*}
    q(y \mid r) &= h(y) \exp\big(T(y) r - A(r) \big)
\end{align*}
for some nonnegative $h$ and $A:\mathbb{F}\to\mathbb{R}$ with $\mathbb{F} \subseteq\mathbb{R}$. For a given $A$, we would like to determine whether there exist an $h$ such that $q(\cdot \mid r)$ defines a valid probability density function (or mass function). This is true if and only if $c h(\cdot)$ for some $c > 0$ is the probability density function (or probability mass function) of a random variable $y$ satisfying
\begin{align*}
    \frac{1}{c} \mathbb{E}_{y \sim c h(\cdot) } \big[ \exp\big(T(y) r - A(r) \big) \big] &= 1 \numberthis \label{eq:valid_pdf}\\
     \mathbb{E}_{y \sim c h(\cdot) } \big[ \exp\big(T(y) r \big)\big] &= c \exp\big( A(r) \big),
\end{align*}
which is the moment generating function of a random variable $T(y)$. Alternatively, we may view $\phi(r) :=  c \exp\big( A(ir) \big)$ as the characteristic function of a random variable $T(y)$ evaluated at real value $r$. Choose $c = \exp\big(-A(0) \big)$. By Bochner's theorem, $\phi$ is a characteristic function if and only if is continuous at $0$ and $\phi$ is positive definite.
\end{proof}
\iffalse
With Lemma~\ref{lem:admissible} at hand, we give an alternative equivalent characterisation of the set of minimal and regular exponential families. Let $\lambda \geq 0$, $\mathbb{F}$ be an open set, and $a:\mathbb{F} \to \mathbb{R}$ be infinitely differentiable such that
\begin{align*}
    A(r) = \lambda \frac{r^2}{2} + a(r)
\end{align*}
is strictly convex, continuous at $r=0$ and $\exp\big(A(i r) \big)$ is positive definite (when viewed as a function of $r$). Then there exists an $h$ and $T$ such that $h, T$ and $A$ form the base pdf, sufficient statistic and log partition function of a regular minimal exponential family. Choosing $\lambda = 0$ recovers the original parameterisation of the set of exponential families. 

The explicit dependence of $A$ on the Gaussian term $\lambda \frac{r^2}{2}$ allows one to slowly ``Gaussianise" otherwise invalid log partition functions by increasing $\lambda$. If $\lambda$ can be increased such that $\exp(A(ir))$ is positive definite, $A$ is an admissible exponential family. 
\fi

\subsection{Antiderivative of hyperbolic tangent}
Choose $A(r) = \int \tanh(r) \, dr = \log \cosh(r)$. One readily observes that $A$ is admissible since $\exp\big( A(ir) \big)= \cos(r)$, which is positive definite.

Since the characteristic function of the base pdf is $2\pi$-periodic, it coincides with a discrete integer-valued random variable. In fact, we can construct an exponential family supported on $\{ -1, 1\}$ with this choice of $A$, $T(y)=y$ and uniform $h(y) = 1/2$ by verifying that~\eqref{eq:valid_pdf} holds with $c=1$. This is a non-interacting Ising model.

\subsection{Rectified canonical parameters}
Appendix~\ref{app:rectifed_gauss} and~\ref{app:rectifed_bernoulli} describe some special cases where $R$ is $\ReLU$ or $\tReLU$.

\subsection{The Bose-Einstein integral}
\label{app:einstein1}
Let 
\begin{align*}
    B_j(r) = \frac{1}{\Gamma(j+1)} \int_0^\infty \frac{y^j}{e^{y - r} - 1} \, dy, \quad j > -1, \quad r < 0
\end{align*}
denote the complete Bose-Einstein integral of order $j$. It is known~\citep{dingle1957bose} that $B_j(r) = \text{Li}_{j+1}\big(e^r\big)$, where $\text{Li}_{j+1}$ denotes the polylogarithm of order $j+1$. The Bose-Einstein integral satisfies a recursive relationship 
\begin{align*}
    \frac{ d }{d r} B_j(r) &= B_{j-1}(r),
\end{align*}
with a closed-form initial value $B_1(r) = -\log( 1- e^r)$.

Define 
\begin{align*}
    p(y \mid r) &= h(y) \exp\big(y r -  B_j(\beta r) \beta^{-(j+1)} \big),
\end{align*}
where $h(y)$ is a nonnegative function and $\beta > 0$ is some fixed value. We would like to determine whether this choice of log partition function is valid. 

 %https://math.stackexchange.com/questions/1607379/equivalence-of-properties-of-characteristic-function-of-a-random-variable
%https://stats.stackexchange.com/questions/475860/characteristic-function-and-fourier-transform-for-a-discrete-random-variable

Since the composition of the exponential function with a positive definite function is positive definite, it suffices to show that the Bose-Einstein integral $ \text{Li}_{j+1}\big(e^{ir} \big)$ is positive definite. This is observed by a comparison to the characteristic function of the Geometric distribution. We have that  
\begin{align*}
    \int_0^\infty \frac{y^j}{e^{y - ir} - 1} \, dy &=  \int_0^\infty \frac{y^je^{-y}e^{i r}}{1 - e^{-y} e^{ir}} \, dy \\
    &=  \int_2^1 \frac{(p-1) \big(-\log(p-1)\big)^j e^{i r}}{1 - (p-1) e^{ir}} \frac{-1}{p-1}\, dp, \qquad p = e^{-y} + 1 \\
    &=  \int_1^2 \Big( \frac{(-1)^j \log(p-1)^j }{p} \Big) \frac{ p e^{i r}}{1 - (p-1) e^{ir}}\, dp.
\end{align*}
Now observe that $\Big(\frac{(-1)^j \log(p-1)^j}{p} \Big)$ is positive in $(1,2)$, and $\frac{1}{1 - (1-p) e^{ir}}$ admits a Fourier series (via the geometric series) $\sum_{k=0}^\infty (p-1)^ke^{i r k}$. Since the Fourier series coefficients are positive, $\frac{1}{1 - (p-1) e^{ir}}$ is positive definite, as is the product of positive definite functions $\frac{ p e^{i r}}{1 - (p-1) e^{ir}}$. The integrand is therefore positive definite, and by L\'{e}vy's continuity theorem the integral is a positive definite function.

Since $\phi$ is $2\pi$-periodic, it is the characteristic function of a discrete random variable taking integer values.

\subsection{A non-example in the Fermi-Dirac integral}
\label{app:einstein2}
The Fermi-Dirac integral is closely related to the Bose-Einstein integral. Let 
\begin{align*}
    B_j(r) = \frac{1}{\Gamma(j+1)} \int_0^\infty \frac{y^j}{e^{y - r} + 1} \, dy, \quad j > -1, \quad r < 0
\end{align*}
denote the complete Fermi-Einstein integral of order $j$. It is known~\citep{dingle1957bose} that $B_j(r) = -\text{Li}_{j+1}\big(-e^r\big)$, where $\text{Li}_{j+1}$ denotes the polylogarithm of order $j+1$. The Fermi-Dirac integral satisfies a recursive relationship 
\begin{align*}
    \frac{ d }{d r} B_j(r) &= B_{j-1}(r) \iff \frac{ d }{d r} -\text{Li}_{j+1}\big(-e^r\big) = -\text{Li}_{j}\big(-e^r\big) \numberthis \label{eq:fermi_dirac_recurse}
\end{align*}
with a closed-form initial value $B_1(r) = \log( 1 + e^r)$.

At first, this might appear to be an attractive way to construct softplus activations and their zero-temperature limits. If $B_2(r)$ were an admissible log partition function, then choosing $R$ to be the identity, $\sigma(r) = A'(r) = \log(1 + e^r)$. However, the Fermi-Dirac integral is not necessarily positive definite. Following the same method as the Bose-Einstein integral, we have
\begin{align*}
    \int_0^\infty \frac{y^j}{e^{y - ir} + 1} \, dy &=  \int_0^\infty \frac{y^je^{-y}e^{i r}}{1 + e^{-y} e^{ir}} \, dy \\
    &=  \int_2^1 \frac{(p-1) \big(-\log(p-1)\big)^j e^{i r}}{1 + (p-1) e^{ir}} \frac{-1}{p-1}\, dp, \qquad p = e^{-y} + 1 \\
    &=  \int_1^2 \Big( \frac{(-1)^j \log(p-1)^j }{p} \Big) \frac{ p e^{i r}}{1 + (p-1) e^{ir}}\, dp.
\end{align*}
As before, observe that $\Big(\frac{(-1)^j \log(p-1)^j}{p} \Big)$ is positive in $(1,2)$. However, $\frac{1}{1 + (1-p) e^{ir}}$ admits a Fourier series (via the geometric series) $\sum_{k=0}^\infty (1-p)^ke^{i r k}$. Since the Fourier series coefficients are \textbf{not always positive}, the Fermi-Dirac integral is indefinite. Hence we cannot conclude that its exponential is a positive definite function.

However, an alternative method of understanding softplus activations is given by~\eqref{eq:recipe}. Choosing a Gaussian likelihood $A(v)=v^2/2$, a softplus activation $\sigma$ implies that
\begin{align*}
    R(\eta) &= \sqrt{\int 2 \log\big(1+\exp(\eta)\big)  \, d\eta } \\
    &= \sqrt{-2 \text{Li}_2(-e^\eta) }. \qquad \text{by~\eqref{eq:fermi_dirac_recurse}}
\end{align*}
Roughly speaking, $\log\big(1+\exp(\eta)\big)$ looks like a smooth ReLU, so its integral $-\text{Li}_2(-e^\eta)$ looks like a squared smooth ReLU. The square root $R(\eta)=\sqrt{-2 \text{Li}_2(-e^\eta) }$ looks like a smooth ReLU. The corresponding soft dropout function looks like a smooth step function, and is given by
\begin{align*}
    \rho(\eta) &= \frac{d}{d\eta} \sqrt{-2 \text{Li}_2(-e^\eta) } \\
    &= \log(1+e^\eta) \frac{1}{\sqrt{-2 \text{Li}_2(-e^\eta)}}.
\end{align*}

\subsection{Logistic sigmoid}
\label{app:sigmoid}
It is well-known that the classical logistic sigmoid function $\sigma(\eta) = \big( 1+\exp(-\eta) \big)^{-1}$ is the inverse-link function, or derivative of the log-partition function, of a Bernoulli exponential family with $A(r)=\log(1+\exp r)$. Under such a model, the logistic sigmoid should be used to model binary-valued data, but it is commonly used in practice for real-valued data. We now provide an alternative method for constructing the logistic sigmoid via a non-linearly parameterised Gaussian, providing a principled derivation for its use in modelling real-valued data. Following the recipe~\eqref{eq:recipe}, we have
\begin{align*}
    R(\eta) &= \sqrt{\int 2 \big( 1+\exp(-\eta) \big)^{-1} \, d\eta} \\
    &=  \sqrt{2 \log(1 + \exp(\eta) \big)} \qquad \text{by~\eqref{eq:fermi_dirac_recurse}}.
\end{align*}
Roughly speaking, this looks like the square root of a smooth ReLU. The corresponding dropout function is small for negative inputs or large inputs,
\begin{align*}
    \rho(\eta) &= \frac{d}{d \eta}  \sqrt{2 \log(1 + \exp(\eta) \big)^{-1} } \\
    &= \sigma(\eta) \frac{1}{\sqrt{2 \log(1+e^\eta)}} .
\end{align*}

\subsection{Bregman divergences}
\label{app:bregman}
In order to describe the effect that the nonlinearity $R$ has on the Bregman divergence, we need to recall some machinery from~\citet{banerjee2005clustering}. In this subsection, without loss of generality, suppose $T$ is the identity. We work with the exponential family form
\begin{align*}
    p(y \mid r) = h(y) \exp\big(y r - A(r) \big),
\end{align*}
which is a probability density function for a scalar-valued random variable. Connections extend to the vector-valued case, but since the setting in our text is restricted to factorised (conditionally independent) scalar-valued densities, we write the connection in terms of scalar-valued objects.

Let $\phi$ be a strictly convex and differentiable function mapping from a convex subset of $\mathbb{R}$ to $\mathbb{R}$. The Bregman divergence $D_\phi$ is defined as
\begin{align*}
    D_\phi(\upsilon_1, \upsilon_2) &= \phi(\upsilon_1) - \phi(\upsilon_2) - (\upsilon_1 - \upsilon_2) \frac{\partial}{\partial \upsilon_2} \phi(\upsilon_2).
\end{align*}
%https://www2.cs.uic.edu/~zhangx/teaching/bregman.pdf
%https://www.dianacai.com/blog/2018/02/16/natural-gradients-mirror-descent/
%https://www.mdpi.com/1099-4300/16/12/6338/htm
% https://arxiv.org/abs/1310.7780
Every Bregman divergence corresponds with exactly one exponential family through a convex conjugate. Concretely, the function $\phi$ that generates the Bregman divergence may be thought of as the convex conjugate of a log partition function $A$,
\begin{align*}
    \phi(\mu) = \sup_{r \in \mathbb{F}} \{ \mu r - A(r) \}.
\end{align*}
We have that~\citep[Theorem 4]{banerjee2005clustering}
\begin{align*}
    p(y \mid r ) = \exp\big( -D_\phi(y, \mu) \big) b_\phi(y),
\end{align*}
where $b_\phi$ is a uniquely determined function that does not depend on the expectation parameter $\mu$ (or the canonical parameter $r$).

The expectation parameter is related to the canonical parameter through $\mu = A'(r)$ and $r = \phi'(\mu)$. In our setting, since the canonical parameter is the result of a canonical map applied to a parameter $\eta$, we have $\mu = A' \circ R \, (\eta)$ and $R(\eta) = \phi'(\mu)$. If $R$ is invertible, then $\eta =\big(  R^{-1} \circ \phi'\big)(\mu)$, otherwise there is more than one value of $\eta$ such that $R(\eta) = \phi'(\mu)$. Finally, note that by definition, $\sigma(\eta) = (A \circ R)' (\eta) = A' \circ R(\eta) \times \rho(\eta)$, so that $\sigma(\eta) / \rho(\eta) = A' \circ R (\eta) = \mu$, whenever the operation  $\sigma(\eta) / \rho(\eta)$ is defined.

As an example, we may obtain the setting of nonlinear least squares by choosing $\phi(\upsilon) = \upsilon^2$. In this case, $A'$ is the identity, so that minimising the negative log likelihood is equivalent to minimising 
\begin{align*}
    D_\phi\big(y, A' \circ R(\eta) \big) &= \Big( y -  R(\eta) \Big)^2.
\end{align*}

More generally, our nonlinearly parameterised exponential family corresponds with a nonlinearly parameterised Bregman divergence $D_\phi\big(y, A' \circ R(\eta) \big)$.

\newpage
\section{The smooth ReLU}
\label{app:relu}
\paragraph{Generalised functions.} Our calculus will involve limits of integrals against measures that converge to the singular Dirac delta measure. In particular, we make use of evaluation and certain derivative properties of the Dirac delta measure using the abuse of notation
\begin{align*}
    \lim_{\tau \to 0} \int_{-\infty}^\infty g(x) \phi_\tau (x+a) \, dx &= \int_{-\infty}^\infty g(x) \delta (x+a) \, dx = \frac{\partial}{\partial a} \int_{-\infty}^\infty g(x) \Theta (x+a) \, dx = g(-a)
\end{align*}
for any continuous and compactly supported $g$, where $\phi_\tau$ is any nascent delta function and $\Theta$ is the Heaviside step function. 

This property will turn out to be useful because we may convert functions that are non-differentiable at a point into differentiable functions by introducing a well-motivated integral and limit. This enables to analyse activation functions such as the rectified linear unit which may be expressed as $\ReLU(\eta) = \Theta(\eta) \eta$.

\paragraph{The $\tReLU$.}
Let $\ReLU(\eta) = \Theta(\eta) \eta $ denote the ReLU, and $\tReLU(\eta) = \int_\mathbb{R} \ReLU(\eta + \epsilon) p_\tau (\epsilon) \, d\epsilon$ denote the $\tau$-smoothed ReLU. Here $p_\tau$ is the PDF of a zero-mean Gaussian random variable with standard deviation $|\tau|$.
We have an alternate representation using the expected value of the absolute value of a Gaussian random variable,
\begin{align*}
    \tReLU(\eta) &= \int \ReLU(\eta + \epsilon) p_\tau (\epsilon) \, d\epsilon \\
    &= \frac{1}{2}\Big( \eta + \mathbb{E}\big[  |\eta + \epsilon | \big] \Big) \\
    &= \eta \Phi\Big(\frac{\eta}{|\tau|} \Big) + \frac{|\tau|}{2} \sqrt{\frac{2}{\pi}} \exp(-\eta^2/2\tau^2),
\end{align*}
where $\Phi$ is the CDF of the standard normal distribution. The first term is also known as the GELU activation function, $\text{GELU}(\eta) = \eta \Phi\big(\frac{\eta}{|\tau|} \big)$. The GELU approaches the ReLU as $\tau \to 0$. The second term is a correction term that approaches $0$ as $\tau \to 0$. Convergence is uniform, since by $1$-Lipschitz of the ReLU,
\begin{align*}
    \big| \tReLU(\eta) - \ReLU(\eta) \big| &= \Big| \int \big( \ReLU(\eta) - \ReLU(\eta + \epsilon) \big) p_\tau (\epsilon) \, d\epsilon \Big| \\
    &\leq  \int | \epsilon | p_\tau (\epsilon) \, d\epsilon \\ 
    &= | \tau | \sqrt{\frac{2}{\pi}}, \numberthis \label{eq:uniform_relu}
\end{align*}
which decreases monotonically as $\tau \to 0$ for all $\eta$. As a convolution with a Gaussian, the $\tReLU$ is infinitely differentiable. The first derivative is
\begin{align*}
    \frac{\partial }{\partial \eta} \tReLU(\eta) &= \frac{\partial }{\partial \eta} \int (\eta + \epsilon) \Theta(\eta + \epsilon) p_\tau (\epsilon) \, d\epsilon \\ 
    &=  \int \Big(\Theta(\eta + \epsilon) +  (\eta + \epsilon) \delta(\eta + \epsilon) \Big) p_\tau (\epsilon) \, d\epsilon \\ 
    &=  \int \Theta(\eta + \epsilon) p_\tau (\epsilon) \, d\epsilon \\ 
    &=  \int_{-\eta}^\infty p_\tau (\epsilon) \, d\epsilon =: \Phi_\tau(\eta)
\end{align*}
which is the CDF of a zero-mean Gaussian random variable with standard deviation $\tau$ evaluated at $\eta$, i.e. $ \Phi_\tau(\eta) = \mathbb{P}(\varepsilon \leq \eta)$, where $\varepsilon \sim \mathcal{N}(0, \tau^2)$. The second derivative is then the PDF of $\varepsilon$ evaluated at $\eta$,
\begin{align*}
    \frac{\partial^2 }{\partial \eta^2} \tReLU(\eta) &=  \frac{1}{\sqrt{2 \pi \tau^2}} \exp\big( -\frac{\eta^2}{2\tau^2}\big).
\end{align*}

\paragraph{Leaky $\tReLU$.}
Let $m \in [0, 1]$ be some gradient parameter. We may define a Leaky $\tReLU$ by
\begin{align*}
    \LtReLU(\eta) &= (1-m) \tReLU(\eta) + m \eta.
\end{align*}
When $m=0$, $\LtReLU$ is $\tReLU$ and when $m=1$, $\LtReLU$ is linear.

\subsection{Rectified positive mean Gaussian}
\label{app:rectifed_gauss}
\begin{figure}[h]
    \centering
    \includegraphics[scale=0.3]{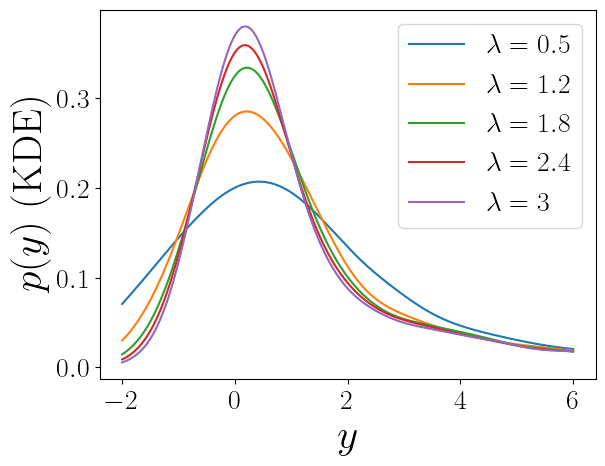}
    \includegraphics[scale=0.3]{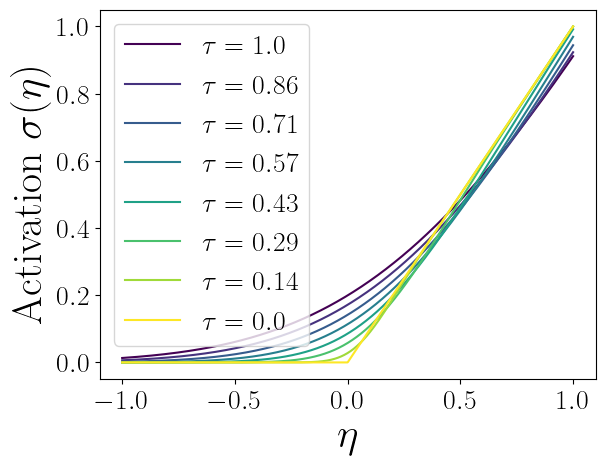}
    \includegraphics[scale=0.3]{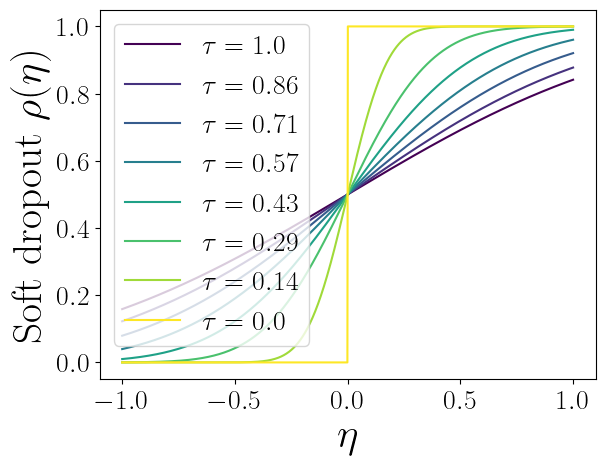}
    \caption{(Left) KDE of hard-rectified positive mean Gaussian marginal likelihood in $1$ dimension with $\tau = 0.5$. (Middle) Activation function $\Phi_\tau(\eta) \tReLU(\eta)$ for different values of $\tau$. (Right) Soft dropout function $\Phi_\tau(\eta)$ for different values of $\tau$.}
    \label{fig:gauss_kde}
\end{figure}

Let $A(r) = r^2/2$ and $R(\eta) = \ReLU(\eta)$. This corresponds with an exponential family likelihood that always has a mean greater than or equal to zero. In order to visualise what this means, consider the case $d=l=1$. With a $\mathcal{N}(0, \sqrt{2})$ prior over $\mathsf{W} \in \mathbb{R}$ and $z \in \mathbb{R}$, the model evidence
\begin{align*}
    p(y) &= \mathbb{E}_{\mathsf{W}, z \sim \mathcal{N}(0,1)} \big[p\big(y \mid \ReLU( \mathsf{W} z) \big)   \big]
\end{align*}
is visualised by sampling $10000$ $\mathsf{W}$, $z$ and $y$ and plotting a KDE for $y$. This is shown in Figure~\ref{fig:gauss_kde}, where the variance of the conditional distribution of $y$ given $\mathsf{W}z$ is $\lambda^{-1}$.

\subsection{Rectified positive logit Bernoulli and binomial}
\label{app:rectifed_bernoulli}
\begin{figure}[h]
    \centering
    \includegraphics[scale=0.3]{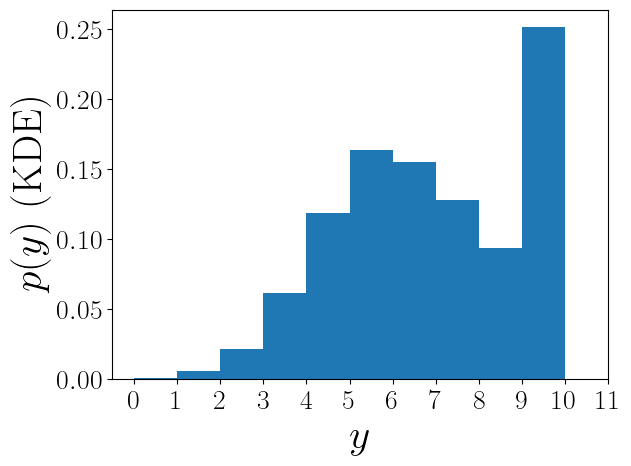}
    \includegraphics[scale=0.3]{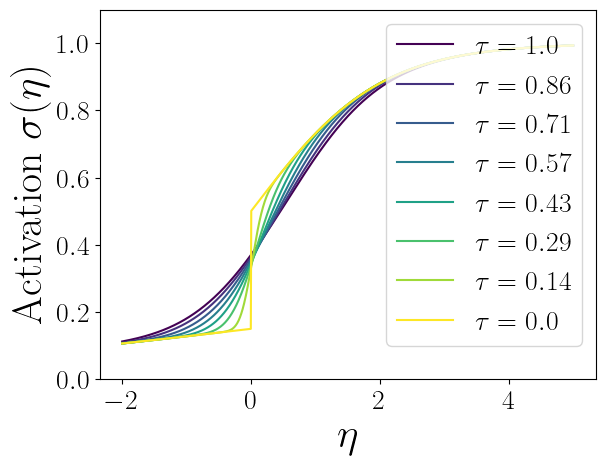}
    \includegraphics[scale=0.3]{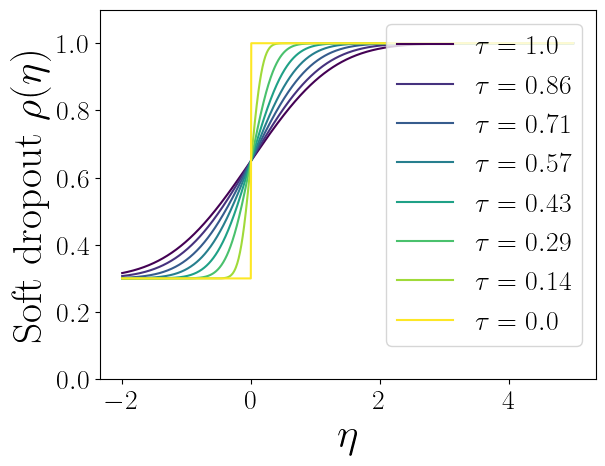}
    \caption{(Left) Histogram of samples from hard-rectified positive logit Binomial marginal likelihood in $1$ dimension with $t=10$ trials and $\tau = 0.5$. The leaky soft-rectified Bernoulli leads to interesting activation functions and dropout functions. Shown are the activation function (middle) and dropout function (right) when $m = 0.3$ for various values of $\tau$. Recall that the activation function is $(A \circ R)'$, which is different to the expectation parameter $A' \circ R$.}
    \label{fig:binomial_hist}
\end{figure}

Let $A(r) = t \log \big(1 + \exp(r) \big)$ and $R(\eta) = \LtReLU(\eta)$. The parameter $t$ is the number of trials for a binomial random variable, and $t=1$ is the special case of a Bernoulli random variable. We may produce similar plots to Figure~\ref{fig:gauss_kde} by plotting histograms of the integrated binomial distribution. The case $t=1$ is not interesting, since this is just a $\{0,1\}$-valued random variable and hence another Bernoulli. The case $t=10$ is shown in Figure~\ref{fig:binomial_hist}.

In this case, the activation function is $\sigma(\eta) = t \frac{\exp(\LtReLU(\eta))}{1+\exp\big(\LtReLU(\eta) \big)} \big( \Phi_\tau(\eta)(1-m) + m \big)$. The soft dropout function is $\rho(\eta) = \Phi_\tau(\eta) (1-m) + m$

\newpage
\section{Proofs}
\label{app:proofs}
\generalshallow*

\begin{proof}
The log posterior over $\mathsf{Z}$ and $\mathsf{W},B$ is
\begin{align*}
    \log p(\mathsf{Z}, \mathsf{W},B \mid \mathsf{Y}) &= \log p(\mathsf{Y} \mid \mathsf{W}, \mathsf{Z},B) + \log p(\mathsf{Z}) + \log p(\mathsf{W},B) - \log p(\mathsf{Y}).
\end{align*}
The joint MAP estimate $(\mathsf{Z}^\ast, \mathsf{W}^\ast, B^\ast)$ solves the constrained optimisation problem
\begin{align*}
    \mathsf{W}^\ast, B^\ast &= \argmin_{W, B \in \mathbb{W}} -\log p\big(\mathsf{Y} \mid R(\mathsf{W} \mathsf{Z}^\ast + B)\big) - \log p(\mathsf{W},B) - \log p(\mathsf{Z}^\ast) \\
    \text{subject to } Z_s^\ast &= \argmin_{Z_s} \frac{\lambda}{2}\Vert Z_s \Vert_2^2 - R (Z_s^\top \mathsf{W}^\top + B^\top)T(Y_s) + 1_{d \times 1}^\top A \circ R \, \big( \mathsf{W}Z_s + B \big).
\end{align*}
The outer problem is converted into a sum by replacing $\log p\big(\mathsf{Y} \mid R(\mathsf{W} \mathsf{Z}^\ast + B)\big)$ with the sum of likelihoods~\eqref{eq:exp_fam} evaluated at the datapoints $Y_s=(y_{si})_{i=1}^d$ and canonical parameters $R(H_s) = R(\mathsf{W} Z_s^\ast + B)$.

We convert the inner DDN layer into a DEQ layer.
We consider the inner problem for fixed $s$, which finds coefficients given $W$ and $b$. 
The stationary points of the inner constraint satisfy
\begin{align*}
    Z^\ast &= \argmin_{z \in \mathbb{R}^l} \frac{\lambda}{2} \Vert z \Vert_2^2 - R (Z^\top \mathsf{W}^\top + B^\top) T(Y) + 1_{d \times 1}^\top A \circ R \,(\mathsf{W} z+B) \\
    0 &= \lambda Z^\ast - \mathsf{W}^\top \big(T(Y) \odot \rho(\mathsf{W}z + B) \big)+ \mathsf{W}^\top \sigma(\mathsf{W} Z^\ast+B) \numberthis \label{eq:deriv_eq_0} \\
    Z^\ast &= \frac{1}{\lambda} \mathsf{W}^\top \big( T(Y) \odot \rho(\mathsf{W}Z^\ast + B)  - \sigma(\mathsf{W} Z^\ast+B) \big). \numberthis \label{eq:deq_fp}
\end{align*}
\iffalse
Suppose $R$ is the identity. Then the problem is strongly convex, since $\lambda >0$ and $A$ is strictly convex. The problem is differentiable everywhere and unconstrained, and as such the first order gradient condition~\eqref{eq:deriv_eq_0} is sufficient to characterise the unique optimum.
This implies that the fixed point equation~\eqref{eq:deq_fp} is guaranteed to admit a unique fixed point.
% https://math.stackexchange.com/questions/1035110/does-every-strongly-convex-function-has-a-stationary-point
%https://math.stackexchange.com/questions/2311335/existence-of-minimizer-for-strongly-convex-function-on-closed-convex-set
% https://www.math.usm.edu/lambers/mat419/lecture4.pdf
\fi
The Hessian $H$ of the constraint is 
\begin{align*}
    H &= \lambda I - \mathsf{W}^\top \Big( \text{diag}\big( T(Y) \odot \rho'(\mathsf{W} Z^\ast + B)  - \sigma'(\mathsf{W} Z^\ast + B) \big) \Big)\mathsf{W} 
\end{align*}
This is positive definite if the largest eigenvalue of $\mathsf{W}^\top \Big( \text{diag}\big( T(Y) \odot \rho'(\mathsf{W} Z^\ast + B)  - \sigma'(\mathsf{W} Z^\ast + B) \big) \Big)\mathsf{W} $ is less than $\lambda$. Let $m$ denote the largest entry of $\text{diag}\big( T(Y) \odot \rho'(\mathsf{W} Z^\ast + B)  - \sigma'(\mathsf{W} Z^\ast + B) \big)$. 

Suppose $m$ is nonnegative. The largest eigenvalue is less than $\lambda$ if $\frac{m}{\lambda}\Vert \mathsf{W}^\top \mathsf{W} \Vert_2 < 1$. Since the objective is strongly convex and continuously differentiable, the solution to the fixed point equation is the unique global minimiser~\citep[Theorem 2.8]{wright_recht_2022}. If $m$ is negative, then the matrix clearly has only positive eigenvalues.

\end{proof}

\relushallow*

\begin{proof}
Any stationary point satisfies
\begin{align*}
    0 &= \frac{\partial}{\partial Z_s^\ast} \Big(\frac{\lambda}{2}\Vert Z_s^\ast \Vert_2^2 - \ReLU ({Z_s^\ast}^\top \mathsf{W}^\top + B^\top)T(Y_s) + 1_{d \times 1}^\top A \circ \ReLU \, \big( \mathsf{W}Z_s^\ast + B \big) \Big) \\
    0 &= \lambda Z_s^\ast - \mathsf{W}^\top T(Y_s) \odot \Theta(\mathsf{W}Z_s^\ast + B) + \mathsf{W}^\top \Theta(\mathsf{W}Z_s^\ast + B) A'\circ \ReLU( \mathsf{W}Z_s^\ast + B )\\
    Z_s^\ast &= \frac{1}{\lambda} \mathsf{W}^\top \big( T(Y_s) \odot \Theta(\mathsf{W}Z_s^\ast + B) - \sigma(\mathsf{W} Z_s^\ast + B) \big),
\end{align*}
and also $WZ_s^\ast + B$ contains no elements that are $0$.
Since $\Theta(\mathsf{W}Z_s^\ast + B) \in \{0, 1\}^d$, there are $2^d$ possible values of $\Theta(\mathsf{W}Z_s^\ast + B)$. Noting that $\sigma(\eta) = \Theta(\eta) A' \circ \ReLU \, (\eta)$, this implies that any $Z_s^\ast$ must be a fixed point solution to one of the $2^d$ equations
\begin{align*}
        Z_s^\ast &= \frac{1}{\lambda} \mathsf{W}^\top \Big( P \odot \big( T(Y_s)  - A' \circ R \, (\mathsf{W} Z_s^\ast + B) \big) \Big)  \numberthis \label{eq:fp_fixed_dropout}
\end{align*}
for some $P \in \{0, 1\}^d$. The right hand side is a contraction whenever $\frac{a}{\lambda}\Vert \mathsf{W}^\top \mathsf{W} \Vert < 1$, since the composition of the two Lispchitz functions $A'$ and $R$ is also Lipschitz. This implies that there exists a unique solution $Z_s^\ast$ to each of the $2^d$ possible fixed point equations. Some of these fixed point equations may not admit solutions such that $P = \Theta(\mathsf{W}Z_s^\ast + B)$, so there are at most $2^d$ fixed points. 

To see that there exists at least one fixed point, construct an iterated function system consisting of $2^d$ functions for each value of $P$. The attractor of such a system is nonempty. In particular for any starting point $Z_s$, we may apply a sequence of maps~\eqref{eq:fp_fixed_dropout} with corresponding values of $P=\Theta(\mathsf{W} Z_s + B)$, which converges to an element of the attractor~\citep[Theorem \S 3.1]{hutchinson1981fractals}.

Now observe that all stationary points are local minima. At the stationary point, where $\mathsf{W} Z_s^\ast + B$ does not contain any zero elements, the Hessian is given by
\begin{align*}
    \lambda I - \mathsf{W}^\top \text{diag}\Big(\Theta \big(\mathsf{W} Z_s^\ast + B \big) A'' \circ R \, \big( \mathsf{W} Z_s^\ast + B\big)\Big) W.
\end{align*}
This matrix is positive definite since $\frac{a}{\lambda}\Vert \mathsf{W}^\top \mathsf{W} \Vert_2 < 1$. Therefore, the stationary points are local minima~\citep[Theorem 2.5]{wright_recht_2022}.
\end{proof}

\newpage
\section{Deep PED}
\label{app:deep}
The joint maximum $({Z^{(1)}}^\ast, \ldots {Z^{(L)}}^\ast, \theta^\ast)$ satisfies for each $l$
\begin{align*}
    {Z^{(l)}}^\ast, {\mathsf{W}^{(l)}}^\ast, {B^{(l)}} &\in \argmax_{Z^{(l)} \in \mathbb{R}^{d_l}, \mathsf{W}^{(l)}, B^{(l)} \in \mathbb{W}^{(l)}} p\big({Z^{(l-1)}}^\ast \mid Z^{(l)}, \mathsf{W}^{(l)} B^{(l)} \big)      p\big(Z^{(l)} \mid {Z^{(l+1)}}^\ast, {\mathsf{W}^{(l+1)}}^\ast, {B^{(l+1)} }^\ast\big)  
\end{align*}
where we understand the second factor to mean $p(Z^{(L)})$ in the case $l=L$.

\paragraph{Stationary points.} Here the conditional distribution of $Z^{(l)}$ given $Z^{(l+1)}$ takes the role of the prior from the shallow network case. 
It behaves like the unconditional Gaussian prior in the sense that its logarithm is quadratic, plus an extra term depending on ${Z^{(l+1)}}^\ast$. 
At a stationary point, each ${Z^{(l)}}^\ast$ must satisfy
\begin{align*}
    0 &=  -{\mathsf{W}^{(l)}}^\top  \Bigg( \rho^{(l)} ( \mathsf{W}^{(l)}Z^{(l)} + B^{(l)} ) \odot   \sqrt{\lambda^{(l-1)}}  {Z^{(l-1)}}^\ast    -  \sigma^{(l)}\big(   \mathsf{W}^{(l)}Z^{(l)} + B^{(l)}  \big)  \Bigg) + \\
    & \lambda^{(l)} Z^{(l)} -  \sqrt{\lambda^{(l)}}  R^{(l+1)}\big(\mathsf{W}^{(l+1)} {Z^{(l+1)}}^\ast +  B^{(l+1)}\big) \\
    {Z^{(l)}}^\ast &= \frac{1}{\lambda^{(l)}}   {\mathsf{W}^{(l)}}^\top  \Bigg( \rho^{(l)} ( \mathsf{W}^{(l)}{Z^{(l)}}^\ast +B^{(l)})  \odot  \sqrt{\lambda^{(l-1)}}  {Z^{(l-1)}}^\ast   -  \sigma^{(l)}\big(   \mathsf{W}^{(l)}{Z^{(l)}}^\ast+B^{(l)}  \big) \Bigg) + \\
    &\frac{1}{\sqrt{\lambda^{(l)}}} R^{(l+1)}\big(\mathsf{W}^{(l+1)} {Z^{(l+1)}}^\ast + B^{(l+1)}\big), \label{eq:fp_one_l} \numberthis
\end{align*}
and additionally $\mathsf{W}^{(l)} {Z^{(l)}}^\ast + B^{(l)}$ must not contain any zero coordinates for any $l$.

\paragraph{Augmenting the state space.}
The fixed point solution for layer $l$ depends on the solutions in layer $l-1$ and $l+1$, ignoring boundary cases. We may jointly compute the fixed points in a single DEQ layer by augmenting the $Z^{(l)}$ variables into a single state of size $D = \sum_{l=1}^L d^{(l)}$. 

Let $\zeta = (Z^{(1)}; \ldots; Z^{(L)} ) \in \mathbb{R}^D$ and write 
\begin{align*}
    G^{(l)}\big( Z^{(l)}; Z^{(l-1)},& Z^{(l+1)} \big) = \frac{1}{\sqrt{\lambda^{(l)}}} R^{(l+1)}\big(\mathsf{W}^{(l+1)} {Z^{(l+1)}} + B^{(l+1)}\big) + \\
    & \frac{1}{\lambda^{(l)}}   {\mathsf{W}^{(l)}}^\top  \Bigg( \rho^{(l)} ( \mathsf{W}^{(l)}{Z^{(l)}} +B^{(l)})  \odot   \sqrt{\lambda^{(l-1)}}  {Z^{(l-1)}}   -  \sigma^{(l)}\big(   \mathsf{W}^{(l)}{Z^{(l)}}+B^{(l)}  \big) \Bigg),
\end{align*}
and
\begin{align*}
    G^{[1:L]}(\zeta) &= \Big( G^{(1)}\big( Z^{(1)}; Z^{(0)}, Z^{(2)} \big); \ldots ;G^{(L)}\big( Z^{(L)}; Z^{(L-1)}, 0 \big)  \Big).
\end{align*}

\paragraph{Counting the stationary points.} If $R$ is $\ReLU$, following the same argument as the proof of Theorem~\ref{thm:relu_shallow}, each $\rho^{(l)}$ maps to a finite set of size $2^{d^{(l)}}$. We may construct an iterated function system with $\prod_{l=1}^L 2^{d^{(l)}}$ elements, with components of the form
\begin{align*}
    G^{(l)}_{P^{(l)}} \big( Z^{(l)}; Z^{(l-1)}, & Z^{(l+1)} \big) = \frac{1}{\sqrt{\lambda^{(l)}}} R^{(l+1)}\big(\mathsf{W}^{(l+1)} {Z^{(l+1)}} + B^{(l+1)}\big) + \\
    & \frac{1}{\lambda^{(l)}}   {\mathsf{W}^{(l)}}^\top  \Bigg( P^{(l)}  \odot   \sqrt{\lambda^{(l-1)}}  {Z^{(l-1)}}   -  \sigma^{(l)}\big(   \mathsf{W}^{(l)}{Z^{(l)}}+B^{(l)}  \big) \Bigg), \label{eq:fp_one_l_ifs} \numberthis
\end{align*}
with a corresponding $G^{[1:L]}_{P^{[1:L]}}(\zeta)$. By the proof of Theorem~\ref{thm:relu_shallow}, at least one $1$ and at most $\prod_{l=1}^L 2^{d^{(l)}}$ stationary points of $G^{[1:L]}(\zeta)$ exist if all of the functions in the IFS are contractions. The Jacobian of $G^{[1:L]}_{P^{[1:L]}}(\zeta)$ is a tridiagonal block matrix. The diagonal block is
\begin{align*}
    -\frac{1}{\lambda^{(l)}} {\mathsf{W}^{(l)}}^\top \text{diag}\Big( {\sigma^{(l)}}'\big(   \mathsf{W}^{(l)}{Z^{(l)}}+B^{(l)}  \big) \big) \Big) \mathsf{W}^{(l)}
\end{align*}
and is always defined, since $\mathsf{W}^{(l)} {Z^{(l)}}^\ast + B^{(l)}$ must not contain any zero coordinates for any $l$. For $l < L$, the right off-diagonal block is
\begin{align*}
    \frac{1}{\sqrt{\lambda^{(l)}}} \text{diag} \Big( {\rho^{(l+1)}}\big(\mathsf{W}^{(l+1)} {Z^{(l+1)}} + B^{(l+1)}\big) \Big) \mathsf{W}^{(l+1)},
\end{align*}
and is similarly always defined.
For $l > 1$, the left off-diagonal block is
\begin{align*}
    \frac{\sqrt{\lambda^{(l-1)}}}{\lambda^{(l)}}   {\mathsf{W}^{(l)}}^\top  \text{diag} \Big( P^{(l)} \Big),
\end{align*}
and is similarly always defined. To show that this tridagonal matrix is a contraction, split the matrix into a sum of a diagonal block matrix, a superdiagonal block matrix, and a subdiagonal matrix then use the subadditive property of matrix norms. The matrix norm of the diagonal matrix is bounded by $\max_l \frac{1}{\lambda^{(l)}} \Vert {\mathsf{W}^{(l)}}^\top \mathsf{W}^{(l)} \Vert$. The matrix norm of the second and third matrices are respectively bounded by $\max_l \frac{1}{\sqrt{\lambda^{(l)}}}\Vert {\mathsf{W}^{(l)}} \Vert$ and $\max_l \frac{\sqrt{\lambda^{(l-1)}}}{\lambda^{(l)}}\Vert {\mathsf{W}^{(l)}}^\top \Vert$.

If $R$ is not $\ReLU$, we may obtain a similar bound involving a factor analagous to Assumption~\ref{ass:w_set}.

\paragraph{Definiteness of Hessian.} To reason about the nature of the stationary points, we must compute the eigenvalues of the Hessian. This Hessian is a block tridiagonal matrix. From the derivative of~\eqref{eq:fp_one_l_ifs} with respect to ${Z^{(l)}}$, the diagonal block of the Hessian is
\begin{align*}
    \lambda^{(l)} \mathsf{I} - {\mathsf{W}^{(l)}}^\top \text{diag}\Big( {\sigma^{(l)}}'\big(   \mathsf{W}^{(l)}{Z^{(l)}}+B^{(l)}  \big) \big) \Big) W^{(l)}
\end{align*}
and is always defined, since $\mathsf{W}^{(l)} {Z^{(l)}}^\ast + B^{(l)}$ must not contain any zero coordinates for any $l$ when $R$ is the $\ReLU$. For $l < L$, the right off-diagonal block is
\begin{align*}
    \sqrt{\lambda^{(l)}} \text{diag} \Big( {\rho^{(l+1)}}\big(\mathsf{W}^{(l+1)} {Z^{(l+1)}} + B^{(l+1)}\big) \Big) \mathsf{W}^{(l+1)},
\end{align*}
and is similarly always defined. For $l > 1$, the left off-diagonal block is
\begin{align*}
    \sqrt{\lambda^{(l-1)}}   {\mathsf{W}^{(l)}}^\top  \text{diag} \Big( \rho^{(l)} ( \mathsf{W}^{(l)}{Z^{(l)}} +B^{(l)})  \Big),
\end{align*}
and is similarly always defined.

Bounding the eigenvalues of this block tridiagonal matrix in a useful way remains an open challenge that we leave for future work.

\newpage
\section{Convolutional layers}
\label{app:conv}

It is well-known that convolutional layers may be represented as sparse matrices with repeated entries. Our results apply when $\mathbb{W}$ is a space of convolutional layers. Forming the equivalent sparse matrix for a convolutional layer is undesirable since it introduces an additional step and requires constraints. Therefore, it is of interest to discuss how to implement PED convolutional layers without forming the equivalent sparse fully connected layer.

The transpose operator can be computed using transposed convolution~\citep{zeiler2010deconvolutional}. In Pytorch, this operator may be implemented using \textit{torch.nn.ConvTranspose2d()}.

The constraints in Assumption~\ref{ass:w_set} and~\ref{ass:w_set_relu} may be computed efficiently when the norm is the spectral norm~\citep{sedghi2018the}. 

We leave practical investigation of convolutional layers for future work.

\newpage
\section{Details of the illustrative example}
\label{app:experiments}

\paragraph{PCA.} We use the scikit learn implementation of PCA with default hyperparameters. We scale the input data into PCA with scikit learn's StandardScaler.

\paragraph{UMAP.} We use the publically available UMAP implementation~\citep[BSD 3-Clause License]{mcinnes2018umap} with default hyperparameters. 

\paragraph{Data-generating process, shallow networks.} We generate a 2D grid of size $10^5 \times 10^5$ with corners $(\pm 5, \pm 5)$. We then remove all points that are not inside the circle of radius $3$ centered at $(2, 2)$, the circle of radius $2$ centered at $(-3, -3)$ or the diamond of side length $1$ centered at $(4, -4)$. These are the ground truth $\mathsf{Z}$, consisting of $42453$ points in the 2D plane. We choose an $A$ and $R$, then repeat the following steps $100$ times to obtain $100$ random datasets. We generate $\mathsf{W}^{(1)} \in \mathbb{R}^{50 \times 2}$ by drawing elements iid from a Gaussian with zero mean and variance $\frac{1}{2}$. We then sample $\mathsf{Y}$ from an exponential family with canonical parameter $R(\mathsf{W} \mathsf{Z})$, log partition function $A$ and sufficient statistic $T$.

\paragraph{Data-generating process, deep networks.} The same as for shallow networks, but we generate $\mathsf{W}^{(l)} \in \mathbb{R}^{d^{(l-1)} \times d^{(l)}}$ from $\mathcal{N}(0, \mathsf{I}/d^{(l)})$for each $l$, with the widths $d^{(l)}$ as described in Figure~\ref{fig:synthetic} and $d^{(0)}=50$ . Each of the $Z^{(l)}$ are from an exponential family with canonical parameter $R(W^{(l+1)} \mathsf{Z}^{(l+1)})$.

\paragraph{PED.} We use PED with the same choice of $A$, $T$ and $R$ that matches the data generating process. This automatically defines the dropout functions (if any) and activation functions in the network. We use a value of $\lambda=0.1$ whenever $R$ is the identity and $\lambda=1$ whenever $R$ is $\ReLU$. The reason for these different choices of $\lambda$ is that PED layers admit a unique fixed point no matter the value of $\lambda$ when $R$ is the identity (see Theorem~\ref{thm:general_shallow}), so $\lambda$ may be small so that the prior is weak. When $R$ is $\ReLU$, larger $\lambda$ help to encourage PED layers to find local minima. We found it important to initialise weights with a small variance, especially for Poisson distributions, where the exponential inverse link function can either overflow or become very large. We use a zero mean iid Gaussian prior over $\mathsf{W}, B$, which is implemented by applying weight decay to the neural network optimiser. We use Adam~\citep{kingma2015adam} to optimise parameters $\mathsf{W}$ and $B$ using default hyperparameters using a batch size of $500$ and weight decay varying with layer (note this is not the same as using weight decay with AdamW, which would not be equivalent to L2 regularisation or Gaussian prior regularisation~\citep{loshchilov2018decoupled}). Weight decay is set to $10L \times d^{(l-1)}$ in layer $l$. For shallow models, we train the network for $30$ epochs, whereas for deep models, we train the network for $10$ epochs, freezing the parameters in layer $l$ for the first $5l$ epochs. We use a publically available DEQ repository~\citep[MIT License]{bai2019deep} to implement our DEQs, with a Anderson acceleration method used as the solver with default hyperparameters.

\paragraph{Downstream task.} We use a small head network Linear(100) -> ReLU -> Linear(1) using default initialisation in Pytorch. The headnet is appended to our pretrained backbone of each of PCA, PED and UMAP and fine-tuned using Adam with default hyperparameters. For PCA and UMAP, which do not contain learnable parameters, the backbone network is fixed. The full network is trained to minimise the sum of squared errors between the output of the network and the function $g(z) = Z_1 + Z_2$, where $Z_1$ and $Z_2$ are respectively the first and second coordinate of $z$. We use a batch size of $500$ and train the network for $200$ epochs. In order to evaluate our model on the downstream task, we turn dropout off. We randomly partition each of the datasets into $80\%-20\%$ training-testing split and evaluate our network on the test set. We report the number of times the network performed the best out of the $100$ random runs for each of PCA, UMAP and PED. In the case of Deep PED, to allow a fair comparison to other techniques, we only use the bottleneck neurons as input to the head network.

\paragraph{Hardware, software and computational cost.}
We parallelised runs over multiple nodes, each consisting of a single Tesla P100-SXM2-16GB GPU. We used CUDA 11.4 and Pytorch 1.11.0. Only PED is GPU-accelerated. We found that typical run times for the dimensionality reduction task were task dependent. Table~\ref{tab:runtimes} shows typical run times for each task.
\begin{table*}
\centering
    \begin{tabular}{l|c|c|c|c}
        & PCA & tSNE & UMAP & PED \\\hline
         Gaussian   & 0.09 & 901.94 & 180.46 & 294.09 \\
         Bernoulli  & 0.11 & 923.85 & 174.71 & 267.81 \\
         Poisson    & 0.10 & 698.34 & 176.33 & 189.87 \\
         $\ReLU$ $L=1$ & 0.10 & 622.25 & 168.11 & 1101.38 \\
         $\ReLU$ $L=2$ & 0.10 & 813.25 & 159.71 & 356.40    
    \end{tabular}
    \caption{Measured run times (seconds) for each dataset and model.}
    \label{tab:runtimes}
\end{table*}

\paragraph{Interpretable neurons.} One interesting feature of deep PED in the case $L > 1$ is that certain collections of neurons have interpretations of latent variables at different layers. For example, if $L=2$ and $d^{(1)}=30$, $d^{(2)} = 2$, $D = d^{(1)} + d^{(2)} = 32$, we obtain deep PED which may be implemented as a single layer of the form~\eqref{eq:deepPED}. Even though this DEQ layer has $32$ neurons, we may pick out the $2$ neurons which represent the latent variables in last layer. For this reason, deep PED might be considered interpretable. We are not aware of any other DEQ layers that possess a similar quality.

\clearpage
\section{Approximation to the posterior}
\label{app:bayesian}
In shallow PED, the Hessian of the Laplace approximation is found by taking the derivative of the right hand side of~\eqref{eq:deriv_eq_0} with respect to $Z^\ast$. We find
\begin{align*}
    H(Z^\ast) := - \log p(Z^\ast \mid \mathsf{Y}, \mathsf{W}, B) &= \lambda \mathsf{I} -  \mathsf{W}^\top \text{diag}\big( T(Y) \odot \rho'(\mathsf{W} Z^\ast + B) -  \sigma'(\mathsf{W} Z^\ast + B)\big)  \mathsf{W}.
\end{align*}
This leads to the Laplace approximation to the posterior,
\begin{align*}
    q(z \mid \mathsf{Y}, \mathsf{W}, B) &= \mathcal{N}\Bigg( z \mid Z^\ast, H(Z^\ast)^{-1} \Bigg).
\end{align*}
Note that the Hessian is positive definite at the local minima that we found, so that inversion of the Hessian is well-defined.

In deep PED, one may work with the expressions found in Appendix~\ref{app:deep}, which lead to a Hessian with sparse block structure. In many applications, only the square block at the final layer is of interest, which leads to
\begin{align*}
    H({Z^{(L)}}^\ast) :&= \lambda \mathsf{I} -  {\mathsf{W}^{(L)}}^\top \text{diag}\big( T(Y) \odot \rho'({\mathsf{W}^{(L)}} {Z^{(L)}}^\ast + B) -  \sigma'({\mathsf{W}^{(L)}} {Z^{(L)}}^\ast)\big)  {\mathsf{W}^{(L)}} \\
    q(Z^{(L)} \mid Y, \theta) &= \mathcal{N}\Bigg( z \mid Z^\ast, H(Z^\ast)^{-1} \Bigg). 
\end{align*}

If the likelihood were linearly parameterised Gaussian, $\sigma'$ would be $1$, $\rho'$ would be zero and we would recover the well-known Gaussian case,
\begin{align*}
    H({Z^{(L)}}^\ast) &= \lambda I + {\mathsf{W}^{(L)}}^\top {\mathsf{W}^{(L)}},
\end{align*}
which is independent of the MAP ${Z^{(L)}}^\ast$.

If the likelihood were nonlinearly parameterised Gaussian with $R=\ReLU$, we observe an interesting dropout effect. Since at the stationary point, $\rho'$ and $\sigma'$ are well-defined (by definition) and are $0$ and $\Theta$ respectively, we find
\begin{align*}
    H({Z^{(L)}}^\ast) &= \lambda I + {\mathsf{W}^{(L)}}^\top \text{diag}\big( \Theta(\mathsf{W}^{(L)} {Z^{(L)}}^\ast + B) \big) {\mathsf{W}^{(L)}}.
\end{align*}
Interestingly, this Hessian only depends on ${Z^{(L)}}^\ast$ through $\Theta(\mathsf{W}^{(L)} {Z^{(L)}}^\ast + B) \big)$.

\end{document}